%% file: main.tex
\documentclass{article}

\usepackage{arxiv}

\usepackage[utf8]{inputenc} % allow utf-8 input
\usepackage[T1]{fontenc}    % use 8-bit T1 fonts
\usepackage{hyperref}       % hyperlinks
\usepackage{url}            % simple URL typesetting
\usepackage{booktabs}       % professional-quality tables
\usepackage{amsfonts}       % blackboard math symbols
\usepackage{nicefrac}       % compact symbols for 1/2, etc.
\usepackage{microtype}      % microtypography
\usepackage{lipsum}		% Can be removed after putting your text content
\usepackage{graphicx}
\usepackage{natbib}
\usepackage{doi}
\input{preamble}

\newtheorem{definition}{Definition}
\title{OG-RAG: Ontology-Grounded Retrieval-Augmented Generation For Large Language Models}

\author{
  Kartik Sharma, Peeyush Kumar, Yunqing Li \\
  Microsoft Research \\
  Seattle \\
}

\begin{document}
\maketitle

\begin{abstract}
This paper presents OG-RAG, an Ontology-Grounded Retrieval Augmented Generation method designed to enhance LLM-generated responses by anchoring retrieval processes in domain-specific ontologies. While LLMs are widely used for tasks like question answering and search, they struggle to adapt to specialized knowledge, such as industrial workflows or knowledge work, without expensive fine-tuning or sub-optimal retrieval methods. Existing retrieval-augmented models, such as RAG, offer improvements but fail to account for structured domain knowledge, leading to suboptimal context generation. Ontologies, which conceptually organize domain knowledge by defining entities and their interrelationships, offer a structured representation to address this gap. OG-RAG constructs a hypergraph representation of domain documents, where each hyperedge encapsulates clusters of factual knowledge grounded using domain-specific ontology. An optimization algorithm then retrieves the minimal set of hyperedges that constructs a precise, conceptually grounded context for the LLM. This method enables efficient retrieval while preserving the complex relationships between entities. OG-RAG applies to domains where fact-based reasoning is essential, particularly in tasks that require workflows or decision-making steps to follow predefined rules and procedures. These include industrial workflows in healthcare, legal, and agricultural sectors, as well as knowledge-driven tasks such as news journalism, investigative research, consulting and more. Our evaluations demonstrate that OG-RAG increases the recall of accurate facts by 55\% and improves response correctness by 40\% across four different LLMs. Additionally, OG-RAG enables 30\% faster attribution of responses to context and boosts fact-based reasoning accuracy by 27\% compared to baseline methods.
\end{abstract}


\maketitle

\input{sec-introduction}

\input{sec-method}
\input{sec-setup}
\input{sec-results}

% \input{sec-discussion}
\input{sec-conclusion}

\bibliographystyle{plainnat}
\bibliography{ref}

\newpage
\input{sec-appendix}

\end{document}

%% file: preamble.tex
\usepackage{xspace}
\usepackage{amsmath}
\usepackage{amsfonts}
\usepackage{mathtools}
\usepackage{xcolor,colortbl}
\usepackage{bm}
\usepackage{algorithm}
\usepackage{algpseudocode}
\usepackage{multirow}
\usepackage{dsfont}
\usepackage{pict2e}
\usepackage{enumitem,kantlipsum}
\usepackage{mdframed} % For framing the question and answer blocks
\usepackage{amsmath}  % For mathematical expressions (if needed)

\newmdenv[
  linecolor=black,
  innerleftmargin=10pt,
  innerrightmargin=10pt,
  innertopmargin=10pt,
  innerbottommargin=10pt,
  backgroundcolor=gray!10,
  skipabove=4pt,  % Space above each block
  skipbelow=4pt   % Space below each block
]{QAblock}

\usepackage{fancybox}
\usepackage{microtype}
\usepackage{graphicx}
\usepackage{booktabs} % for professional tables

\usepackage{hyperref}

\usepackage{subcaption}
\usepackage{makecell}
\usepackage{wrapfig}
\usepackage{balance}
\usepackage{lipsum}

\usepackage{adjustbox} % For adjustbox environment
\usepackage{geometry}
\usepackage[edges]{forest}    % For tree structures
\usepackage{tikz}      % For drawing (dependency for forest)
\usepackage{xcolor}    % For colors in the tree (optional)
\usepackage{hyperref}  % For clickable URLs

\usetikzlibrary{shapes.geometric, arrows.meta} % Additional TikZ libraries (optional)

\newcolumntype{H}{>{\setbox0=\hbox\bgroup}c<{\egroup}@{}}

\DeclareMathOperator*{\argmax}{arg\,max}

\DeclareMathOperator*{\topkargmax}{top-k\,arg\,max}

\newcommand{\ie}{\textit{i.e.}\xspace}

\newcommand{\Zbf}{\mathbf{Z}\xspace}

\newcommand{\CD}{\mathcal{D}\xspace}

\newcommand{\CO}{\mathcal{O}\xspace}

\newcommand{\CV}{\mathcal{V}\xspace}
\newcommand{\CE}{\mathcal{E}\xspace}
\newcommand{\CA}{\mathcal{A}\xspace}

\newcommand{\CM}{\mathcal{M}\xspace}
\newcommand{\CC}{\mathcal{C}\xspace}

\newcommand{\CI}{\mathcal{I}\xspace}
\newcommand{\CP}{\mathcal{P}\xspace}
\newcommand{\CS}{\mathcal{S}\xspace}

\newcommand{\CH}{\mathcal{H}\xspace}
\newcommand{\CN}{\mathcal{N}\xspace}

%% file: sec-introduction.tex
\section{Introduction}\label{sec:introduction}
\begin{figure}[t]
    \centering
    \includegraphics[width=1.0\linewidth]{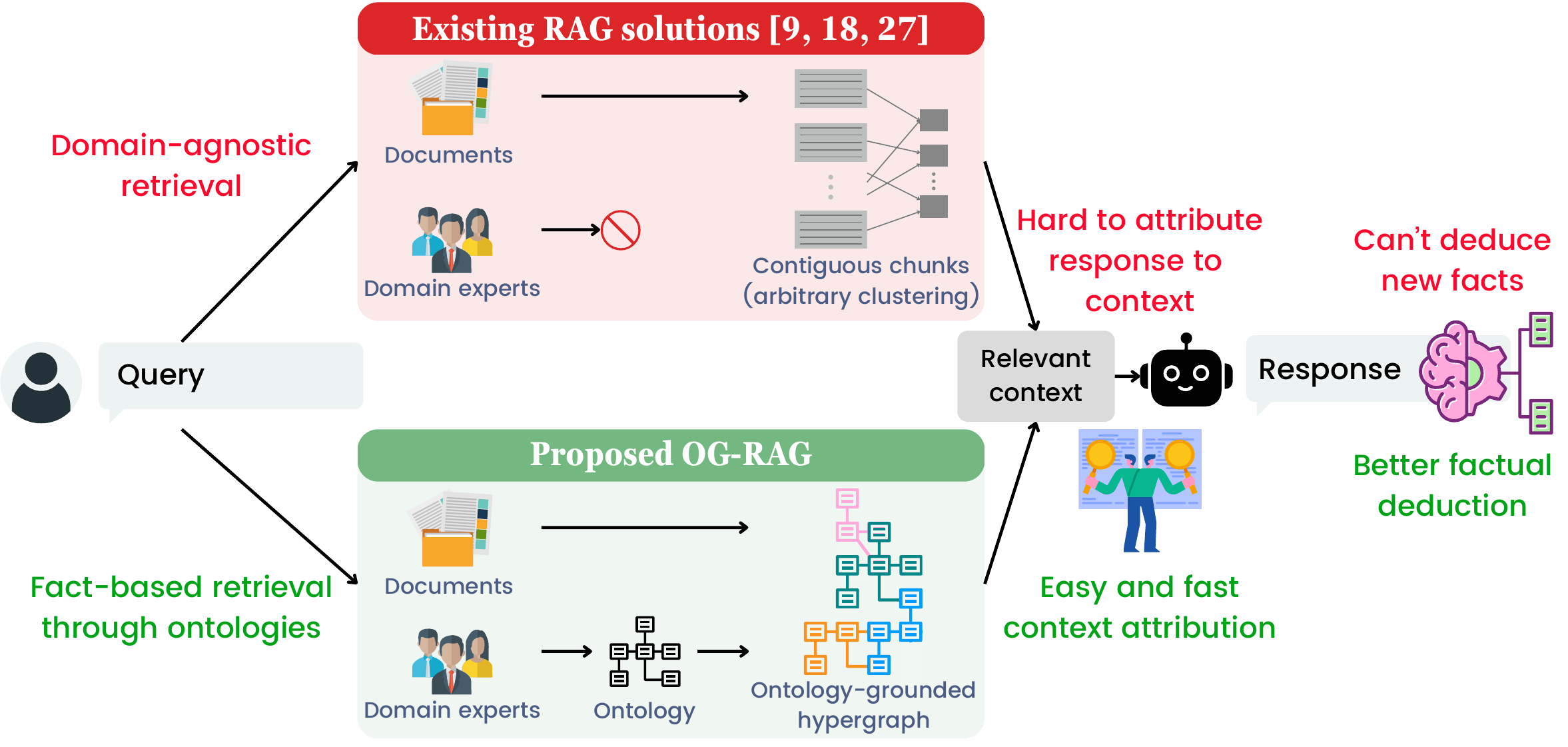}
    \caption{Comparison of the proposed OG-RAG with existing retrieval-augmented generation (RAG) solutions.
    % \hlc{Condense the points for RAG and OG-RAG into a single line and remove the middle icons} \ks{add citations and table, make it more concise}\hlc{Also where is this figure referenced in the main text?}
    }
    \label{fig:motivation}
\end{figure}
Large language models (LLMs) have advanced the capabilities of question-answering systems, search engines, and task-oriented chatbots~\citep{perplexity,chatgpt,achiam2023gpt}. However, they face significant challenges with fact-based adaptation, particularly in domains that rely on precise, domain-specific data~\citep{cascella2023evaluating,thirunavukarasu2023medicine,singhal2023large,guha2024legalbench,wang2024can,balaguer2024rag}. Consider a precision agriculture system where real-time changes in soil moisture and weather data must influence irrigation decisions. A general-purpose LLM might suggest irrigation plans based on broad knowledge but fail to account for specific soil conditions or plant requirements in that region. This lack of adaptability means the LLM's recommendation could be inaccurate, potentially leading to overwatering or under-irrigation, which can harm crops. Such scenarios highlight a core limitation: the inability of LLMs to reliably adapt to domain-specific decision-making, where accuracy and specialized knowledge are paramount~\citep{rudin2019stop,sharma2020machine}.

To overcome these limitations, off-the-shelf LLMs can be either fine-tuned for specific domains~\citep{bommasani2021opportunities} or paired with external tools or documents~\citep{lewis2020retrieval,zhuang2023toolqa,schick2024toolformer}. However, fine-tuning is computationally expensive and requires extensive data curation, making it a less practical solution~\citep{balaguer2024rag,ovadia2023fine}. On the other hand, retrieval-based approaches, such as RAG~\cite{lewis2020retrieval,sarthi2024raptor,zhang2024raft,borgeaud2022improving,karpukhin2020dense,edge2024graphrag}, use domain-agnostic embeddings to retrieve query-relevant information from domain-specific documents and use the retrieved information for answering. Although promising, these methods fail to capture the deep conceptual relationships and nuanced facts required for accurate domain-specific retrieval. 

Each domain organizes its knowledge and terminology in distinct ways, which cannot be generalized across different fields~\citep{mernik2005domain}.  For example, in industrial workflows, facts and relationships are carefully curated and structured into domain-specific frameworks, while in knowledge work and investigative research, ontologies serve as templates for organizing and analyzing facts and concepts~\citep{jackson1990introduction,guarino2009ontology}. Current LLMs struggle to adapt to these diverse structures, limiting their accuracy and effectiveness in specialized domains. Another major issue is that users often struggle to trace generated responses back to the relevant context. Furthermore, many specialized domains follow strict procedural rules, and the current techniques fail to reliably deduce accurate conclusions based on this established domain knowledge. This gap presents a major challenge to the wider applicability of LLMs in specialized workflows.

In this paper, we present OG-RAG (Ontology-Grounded Retrieval Augmented Generation) that bridges this gap of existing LLMs by integrating domain-specific ontologies for fact-based adaptation. Ontologies, which define key entities and their relationships within a domain, provide structured representation that is essential for adapting to complex and evolving information landscapes. As shown in Figure~\ref{fig:motivation}, OG-RAG leverages these ontologies to enhance LLM responses by grounding retrieval within structured domain knowledge, leading to \textit{improved response accuracy}, supporting \textit{flexible fact-based adaptation}, and enabling \textit{verifiable context attribution}. A key feature of OG-RAG is its use of hypergraph representations of domain documents, which is a more sophisticated and multi-faceted way to model relationships than traditional retrieval approaches. Each hyperedge in the hypergraph represents a collection of related factual knowledge based on the corresponding ontology. Using a greedy algorithm, the engine retrieves a minimal set of hyperedges for a given query that forms a compact context for the LLM. Unlike traditional retrieval approaches, this method effectively distills complex relationships and domain-specific knowledge into a structured context, adapting LLMs to generate context-aware responses without adding significant computational overhead. 

OG-RAG applies to a wide set of domains where fact-based adaptation is essential. These include industrial workflows in healthcare, legal, and agricultural sectors, as well as knowledge work such as news journalism, web based investigative research, consulting, and more. Our evaluations of OG-RAG within the agriculture and news domains demonstrate that OG-RAG increases the recall of accurate facts by $55$\%, and improves the overall correctness of generated responses by $40$\% across four different LLMs. Our user study shows that attributing LLM responses to the context retrieved by OG-RAG is $30\%$ faster and better. Finally, in a fact-based reasoning task, we find that LLM responses are $27\%$ more correct when applying pre-defined rules over OG-RAG's context compared to other methods. These results highlight OG-RAG's effectiveness in providing more reliable, fact-based answers in specialized workflows.

\input{sec-relatedWork}

\section{Summary of Key Contributions}
This paper presents several key contributions that address the challenges of fact-based reasoning and hallucination reduction in large language models (LLMs):

\begin{itemize}[leftmargin=*]
    \item \textbf{Fact-based Context Retrieval Enabling Domain Adaptation}: OG-RAG is a novel framework that integrates domain-specific ontologies into the LLM retrieval process, facilitating precise, fact-based adaptation across domains. OG-RAG achieves this through two core mechanisms: (a) formalization of \textit{facts} by transforming data-mapped ontologies into a hypergraph, and (b) using an optimization-based hypergraph retrieval to extract compact clusters of facts that precisely align with domain-specific queries, resulting in more contextually accurate responses.
    \item \textbf{Factual Deduction}: OG-RAG enables factual deduction by leveraging domain-adapted facts to infer conclusions based on established knowledge and relationships. This enables the system to generate accurate, domain-specific conclusions dynamically, improving reasoning over complex factual contexts and enhancing the precision of LLM responses.
    
    \item \textbf{Improved Context Attribution}: OG-RAG enhances response verification by providing precise, fact-based context attribution. This enables both humans and machines to accurately trace conclusions back to their underlying facts, ensuring reliable and transparent fact-checking in domain-specific queries.
    
\end{itemize}
Section~\ref{sec:results} presents extensive evaluations to support each of the enumerated contributions.

%% file: sec-relatedWork.tex
\section{Related Work}

\textit{\textbf{Fine-tuning.}} One approach to overcome the limitations of LLMs is fine-tuning on domain-specific data~\citep{bommasani2021opportunities}. Fine-tuning allows models to adapt to the nuances of a specific domain by retraining the model on specialized datasets. However, this method is computationally expensive, requiring significant resources and extensive data curation, which makes it impractical for many real-world applications~\citep{kumar2022fine}. OG-RAG addresses this shortcoming by eliminating the need for costly fine-tuning through retrieval-based solutions.

\textit{\textbf{Hallucination mitigation.}} LLMs are prone to generating hallucinations, \ie, outputs that are factually incorrect or irrelevant to the input~\citep{ji2023hallucination}. These hallucinations are especially problematic in domains that require precision, such as scientific research or industrial workflows~\citep{thirunavukarasu2023medicine}. Existing systems have attempted to mitigate hallucinations through post-generation correction methods and factuality checks, but these often require additional layers of computation and are not foolproof~\citep{ji2023towards,madaan2024self,welleck2024decoding}. OG-RAG reduces hallucinations by transforming data-mapped ontologies into hypergraphs and uses optimized retrieval of relevant fact clusters, ensuring LLM responses are grounded in domain-specific facts.

\textit{\textbf{Retrieval methods}}
In addition to the traditional retrieval augmented generation (RAG)~\citep{lewis2020retrieval},
graph-based approaches have also been proposed. These include GraphRAG~\citep{edge2024graphrag}, RAPTOR~\citep{sarthi2024raptor}, and other knowledge graph-based frameworks such as Langchain~\citep{langchainrag} and Neo4J~\citep{neo4jUsingKnowledge}. They have advanced LLM performance by leveraging structured knowledge graphs to organize and retrieve contextually relevant information. GraphRAG excels in semantic clustering by organizing entities and relationships, allowing for more efficient handling of complex queries, while RAPTOR uses a hierarchical structure for multi-level abstraction, improving contextual understanding across large documents. However, these approaches rely on ad-hoc extraction of entities and domain-specific information, often without grounding in domain expertise. This results in overly complex workflows for generating the correct structured representation, while still leaving significant gaps in precision. It also leads to weaker context attribution, making it more difficult to trace conclusions back to relevant facts. Improving on these approaches, OG-RAG’s hyperedge construction offers a compact fact representation that enhances transparency through better context attribution, while its hypergraph retrieval mechanism selects optimal fact clusters precisely tailored to the query.

\textit{\textbf{Attribution.}} To enhance the interpretability and reliability of the LLM responses, it is important to attribute their generation to trustworthy sources. One way is to generate text with citations but prior work has shown limitations of existing zero-shot approaches~\citep{gao2023enabling} and specially-trained models~\citep{khalifa2024source}. Furthermore, other forms of attribution are also explored since citations require users to search over a full page to verify the claims in the generated response, which is undesirable. Thus, locally-attributable methods~\citep{slobodkin2024attribute} and human-in-the-loop~\cite{kamalloo2023hagrid} strategies have also been proposed. While these approaches provide sentence-level attribution, complementary benefits can be achieved through interpretable RAG contexts. OG-RAG provides easy-to-attribute contexts that require only a little effort from the users to trace the generation of the response. 

% zero-shot LLMs are shown to be  in providing accurate citations, motivating specific retraining strategies~\citep{khalifa2024source}.
% especially when the attribution
% help in attributing the LLM g are  to attribute their generation to trustworthy sources of data, \eg, legal documents~\citep{}, websites~\citep{}, papers~\citep{}, etc. 

% ~\citep{gao2023enabling}, ~\citep{slobodkin2024attribute}, ~\citep{kamalloo2023hagrid}, ~\citep{khalifa2024source}

\textit{\textbf{Deductive reasoning.}}
Traditional rule-based reasoning systems provide interpretable and easily controllable ways to deduce novel conclusions from a given input~\citep{jackson1990introduction,saparov2023testing}. However, they lack the flexibility and generalization capabilities of neural models like LLMs. On the other hand, LLMs are prone to arbitrary hallucinations in deductive reasoning, which can be problematic in structured workflows~\citep{wang2024can,saparov2023testing}. OG-RAG combines the structured precision of fact-based reasoning with neural flexibility by anchoring unstructured text to domain-specific vocabulary, enabling LLMs to more effectively apply domain-specific rules while maintaining scalability across multiple domains.

%% file: sec-method.tex
\section{OG-RAG}\label{sec:method}
\begin{figure*}[t]
    \centering
    \includegraphics[width=0.9\linewidth]{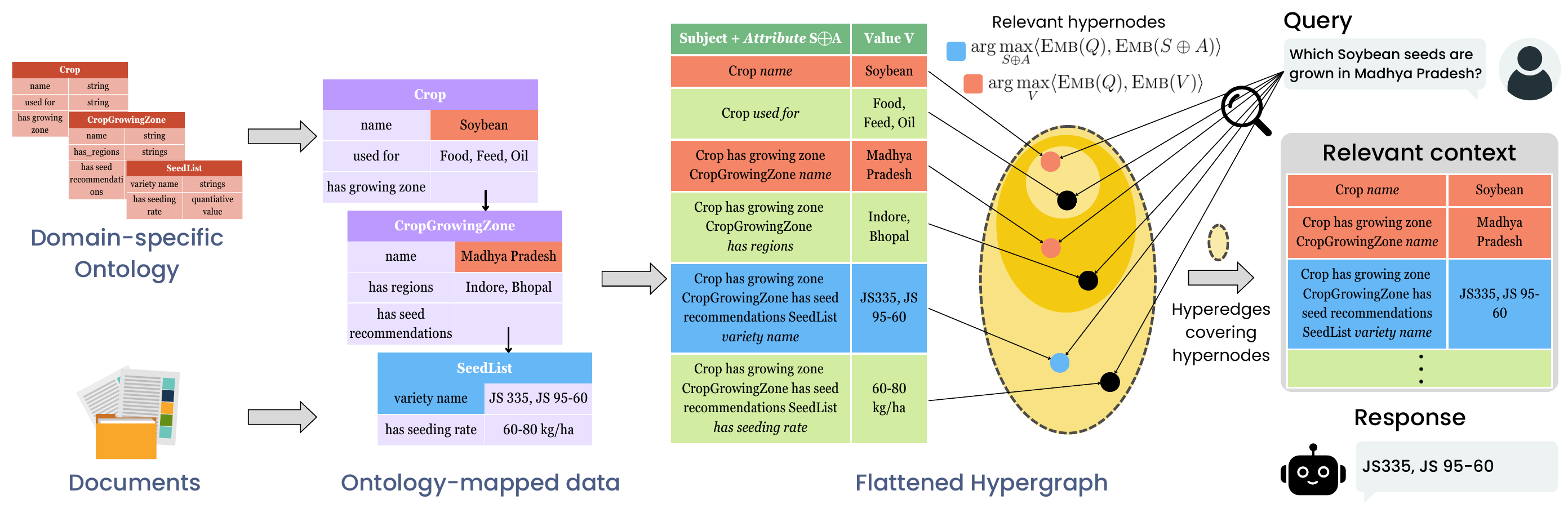}
    \caption{OG-RAG: Ontology-Grounded Retrieval-Augmented Generation
    % \hlc{Follow a particular query so that we show how the contributions in this paper are reflected in OG-RAG}
    }
    \label{fig:pipeline}
\end{figure*}

Ontology Grounded Retrieval Augmented Generation (OG-RAG) is a novel framework which integrates ontologies—formal representations of domain-specific concepts and their relationships—into the retrieval process. Unlike existing retrieval-augmented generation (RAG) systems or other ontology based approaches, which rely on general-purpose embeddings or ad-hoc context generation without grounding in domain expertise, OG-RAG leverages ontology-driven hypergraph retrieval to dynamically adapt LLMs to structured knowledge bases and complex domain-specific queries. Figure~\ref{fig:pipeline} shows the high-level pipeline of the proposed method while we describe each component in more detail below.

\subsection{Hypergraph Construction}

The first part includes mapping the general domain-specific documents $\CD$ onto a given ontology $\CO$ and converting the available information into a usable format for retrieval.

\subsubsection{Ontology}\label{sec:background_ontology}
Different domains organize decision-making by following specific rules and procedures tailored to their unique workflows. In agriculture, for instance, factual information is organized differently depending on the task. Crop cultivation relies on facts like soil quality, weather patterns, and pest management strategies, where decisions follow a series of steps based on phenotype and environmental data. On the other hand, agricultural budgeting uses the same foundational facts—such as farm size or crop type—but applies them through financial models and cost projections, which require a different decision-making framework.

An \textbf{ontology} is a formal representation of key entities and their relationships within a domain. For example, in the agriculture domain, entities like crops, soil, and weather conditions are defined, along with relationships such as "crop is grown in a region" or "soil has moisture level." By defining these entities and relationships, the ontology provides a consistent and clear framework for organizing domain knowledge~\citep{guarino2009ontology,jackson1990introduction}. It differs from \textit{taxonomy} or \textit{classifications} as it allows for richer relationships between entities that need not be hierarchical. More formally, 

\begin{definition}[Ontology]
An \textbf{ontology} \( \CO \subseteq \CS \times \CA \times (\CS \cup \{\phi\}) \) consists of a set of triples that relate a set of entities \( \CS \) using a set of attributes \( \CA \), where \( (s, a, v) \in \CO \) denotes that the subject entity \( s \) has an attribute \( a \), and the value \( v := v_{\CO}(s, a) \) is either:
\begin{itemize}
    \item Another entity \( s' \in \CS \), or
    \item An unspecified domain value, denoted by \( \phi \).
\end{itemize}
Here, \( v := v_{\CO}(s, a) \) represents the value of the attribute \( a \) for entity \( s \), which is either another entity within the ontology or an undefined (unspecified) text or data.
\end{definition}

For example, consider a subject entity $s=$ \textcolor{blue}{``Crop''}, that can have the attribute $a_1=$ \textcolor{teal}{``is grown in''}, which maps it to another object entity $v_{\CO}(s, a_1) = s'=$ \textcolor{blue}{``Crop Region''}. Additionally, the same entity $s$ can have another attribute $a_2=$ \textcolor{teal}{``has name''}, which maps it to an arbitrary text, denoted as $v_{\CO}(s, a_2)=\phi$, indicating that this value is unspecified and can be any relevant text or name in the domain.

Extracting factual information from domain-specific documents $\CD$ is challenging due to their specialized language and often underspecified structure. Moreover, relevant facts are frequently scattered across various unrelated documents. An ontology $\CO$ provides a structured way to organize key terms and their relationships within a domain.

To address this, we propose leveraging the explicit relationships defined in the ontology to extract factual information from these documents. We claim that since domain-specific facts are often grounded in the underlying ontology, enforcing these relationships can help enrich and disambiguate the information contained in the documents. In particular, we use the documents to find values for attributes by extracting relevant domain-specific text or values from the documents themselves (\ie, when $v_{\CO}(s, a) = \phi$). Since domain-specific documents may contain a variety of facts, this value assignment does not have to be unique across all documents. Instead, different parts of the documents may provide distinct yet valid text/data value related to the same ontology entity.

Therefore, we model the extracted information $\CI:=\CD(\CO)$ using a set of self-contained \textit{factual-blocks} $F \in \CD(\CO)$, where each \textit{factual-block} $F$ consists of relationships that map ontology entities to either a unspecfied domain text $\in \CV$ or another entity within the same factual-block $F$. More formally,
\begin{definition}[Ontology-mapped data]\label{defn:map}
    Ontology-mapped data $\CI:=\CD(\CO)$ is information derived from the documents $\CD$ using the ontology $\CO$. It consists of a set of factual-blocks, where each factual-block $F$ represents a set of ontology relationships. For any relationship $(s, a, v) \in F$, the value $v$ is derived as follows: If value $v_{\CO}(s, a) = \phi$ then $v \in \CV$ is extracted from the document text; otherwise $v = v_{\CO}(s, a)$ is the value provided by the ontology. 
\end{definition}
Thus, ontology-mapped data represents \textit{self-contained} and \textit{ontology-grounded} information extracted from domain-specific documents. For example, a factual-block $F$ might represent that: a term $s=$\textcolor{blue}{``Seed''} is $a_1=$ \textcolor{teal}{``of crop''} $v(s, a_1)=$\textcolor{magenta}{``Soybean''} is $a_2=$ \textcolor{teal}{``is grown in''} $v(s, a_2)=$ \textcolor{magenta}{(} $s'=$\textcolor{blue}{``Crop Region''}, which $a_3=$ \textcolor{teal}{``has a name''} of $v(s', a_3)=$ \textcolor{magenta}{``Northwest Region''} \textcolor{magenta}{)}. 

We can apply various pattern-matching heuristics, rule-based strategies, or embedding similarity to generate this ontology-mapped data~\citep{otero2015ontology,jackson1990introduction}. However, with the powerful capabilities of LLMs, we leverage their natural language understanding capabilities to map these ontology entities to corresponding document text more effectively~\citep{babaei2023llms4ol}. We perform this task by prompting the LLM to generate the mapped information in JSON-LD format. The complete prompt is provided in Appendix~\ref{app:ontology_prompt}.

One limitation of this method is that domain-specific ontologies may not always be available or sufficiently comprehensive.  To address this, we are developing an \textit{ontology learning} method that can automatically generate a robust baseline ontology. This provides domain experts with a starting point, making it easier for them to edit and refine an existing ontology rather than building one from scratch. Additionally, in many fields, rich pre-existing ontologies are already available due to decades of research in data modeling and ontology development, which can be directly leveraged by this method. The details of this work are beyond the scope of the current paper, so we omit them.

\subsubsection{Hypergraph Transformation}
Due to the nested structure of the definitions in the factual-block $F \in \CD(\CO)$, directly processing the information in these blocks is challenging. The combinatorial nature of the multi-layered relationships and dependencies make it difficult to efficiently extract or attribute information, which hampers our goal of providing compact and accurate context attribution. To address this, we flatten the structure so that each factual-block $F$ in the ontology-mapped data $\CI$ is converted into a set of flattened factual-blocks $\overline{F}$, making the information easier to handle without significant loss of detail. Algorithm~\ref{alg:flatten} outlines the flattening process, which is also illustrated in Figure~\ref{fig:pipeline}.

\begin{algorithm}[t]
    \caption{Flattening a factual block}
    \label{alg:flatten}
    \begin{algorithmic}[1]
        \Require{Factual block $F$, Concatenation operator $\oplus$.}
        \Ensure{A set of flattened factual-blocks $\overline{F} \gets \, $\Call{Flatten}{$F$} flattens any nested information present in $F$.}
        \Procedure{Flatten}{$F$}
            \State $\overline{F} \gets \{\}$
            % \For{$F \in X$}
            \State $\overline{F}_0 
            \gets \{(s\oplus a, v): (s, a, v) \in F, v \in \CV, (s', a', s) \notin F\}$. \Comment{no dependencies, can be directly flattened}
            \State $\overline{F} \gets \overline{F} \cup \{\overline{F}_0\}$.
            \For{$(s, a, s') \in F \backslash \overline{F}_0$}.
                \If{$s' \in \CS$} % \notin \CV$}
                    \State $F_{s'} \gets \overline{F}_0 \cup \{(s \oplus a \oplus s' \oplus a', v'): (s', a', v') \in F\}$.
                    \State $\overline{F} \gets \overline{F} \;  \cup \, $ \Call{Flatten}{$F_{s'}$}. \Comment{flatten nesting of $s'$}
                    \EndIf
                \EndFor 
            % \EndFor
            \State \Return{$\overline{F}$}
        \EndProcedure
    \end{algorithmic}
\end{algorithm}

We define each flattened factual-block as a hyperedge $e \in \CE$, where a hyperedge $e$ connects multiple hypernodes $\{n_i \in \CN\}$, where each hypernode $n_i \in \overline{F}$ is a primitive set in the flattened-block that can be represented as a key-value pair. Importantly, this flattening process maintains the integrity of the information without introducing data loss. This allows OG-RAG to capture multi-dimensional relationships between facts, unlike simpler graph-based models that only handle pairwise connections. We can now convert the extracted information from the ontology into a more structured hypergraph, defined as follows:
\begin{definition}[Hypergraph]\label{defn:hypergraph}
    A hypergraph $\CH:=(\CN, \CE)$ consists of hypernodes $\CN$ and hyperedges $\CE$, such that each hyperedge $e \in \CE$ is a set of nodes with arbitrary length. Defining $\CP(X)$ as the power set of $X$ and $\bigoplus X$ as the set that is formed by concatenating the strings within each element of the set $X$, we have the hyperedges $\CE \subseteq \CP({\CN})$ and the hypernodes $\CN \subseteq \left[\bigoplus\CP(\CS\times\CA )\right]\times\CV$, where $\times$ is the cartesian product.
\end{definition}
Thus, the set of all flattened factual-blocks extracted from the ontology-mapped documents $\overline{\CI}$ thus can be seen as a hypergraph. We call this simply $\CH(\CI)$. With this definition, a hypernode is essentially a key-value pair and we declare a hyperedge to be a true \textit{fact} grounded in domain-specific data. Mathematically, 

\begin{definition}[Fact]\label{defn:fact}
    A \textit{fact} is a logical assertion between two entities - subject and object, through a functional attribute, which can be evidentially verified to be either true or false. Formally, it can be expressed as a logical assertion that can be verified to have a value of True or False. For example, consider the assertion: hasCropYield(Farm A) $= 500$ tons, where \textit{hasCropYield} is the functional attribute mapping a \textit{farm} (subject) to a \textit{crop yield} (value), and which can be evidentially verified to be either True or False. 
\end{definition}

Therefore, in OG-RAG a hyperedge can be viewed as a representation of a complex fact. Without loss of generality, consider two hypernodes, $n_1(s_1\oplus a_1, v_1) = $(Crop has name, Soybean) and $n_2(p_2 \in \bigoplus\CP(\CS\times \CA), v_2) = $(Crop has growing zone CropGrowingZone with name, Northwest) forming an hyperedge $e = $((Crop has name, Soybean), (Crop has growing zone CropGrowingZone with name, Northwest)) can be represented as a simplified fact:
\begin{center}
\textit{hasGrowingZone}(Crop has name Soybean) = \textit{Northwest},
\end{center} which can be evidentially verified to be True or False.

In this way, the OG-RAG hypergraph construction enables a compact and accurate representation of \textit{facts} that are adapted to the specific domain. This structure facilitates fact verification by allowing users to inspect the hyperedges, which encapsulate the relationships and dependencies between entities.

\subsection{Hypergraph-based retrieval}
With the hypergraph constructed on domain-specific information, \ie, $\CH(\CI(\CD, \CO))$, OG-RAG is now ready to retrieve relevant context based on user query $Q$ that can support the LLM in generating accurate, domain-specific responses.

\subsubsection{Relevant Nodes}
We first identify the set of hypernodes relevant to a given query. Using Definition~\ref{defn:hypergraph}, a hypernode $n \in \CN$ can be represented as a key-value pair that comes from the elements in the sets $\CS$, $\CA$, $\CV$. A hypernode can then be considered relevant to a query if: (1) the query pertains to an attribute $a$ of the term $s$, or (2) the query focuses on an object with specific values $v$.  In other words, a hypernode is relevant if either the similarity between the key (representing concatenated entities and attributes) and the query $Q$ is high, or the similarity between $v$ (the value) and the query $Q$ is high. OG-RAG finds two sets of query-relevant hypernodes: $\CN_S(Q)$ and $\CN_V(Q)$ to represent the two sets respectively. In particular, $\CN_S(Q)$ denotes the top $k$ hypernodes with the highest similarity between their attributed term, \i.e., $s \oplus a$ and the query $Q$ in the vector space $\Zbf$. Similarly, $\CN_V(Q)$ represents the top $k$ hypernodes with the highest similarity between their value $v$ and the query $Q$. Thus, for each query, we extract $2\cdot k$ relevant hypernodes.

\begin{algorithm}[t]
    \caption{Ontology-grounded Retrieval Augmented Generation}
    \label{alg:ograg}
    \begin{algorithmic}[1]
        \Require{Query $Q$, Domain-specific Ontology $\CO$, Documents $\CD$, Sentence embedding function $\Zbf$, LLM $\CM_0$, Maximum length $L$}
        \Ensure{Retrieved context $\CC_{\CH}(Q)$ is grounded in the ontology and relevant to the query}
        \Procedure{OG-Preprocess}{$\CO, \CD, \CM_0$}
            \State $\CI \gets $\Call{LLM $\CM_0$}{Ontology Map ($\CD, \CO$} \Comment{See Definition~\ref{defn:map}}
            \State $\CH(\CI) \gets$ Hypergraph with edges $\bigcup_{F \in \CI}\Call{Flatten}{F}$.
        \EndProcedure

        \Procedure{OG-Retrieve}{$Q, \CH(\CI), \Zbf, k, L$}
            \State $\CN, \CE \gets $  nodes and edges of the hypergraph $\CH(\CI)$.
            \State $\CN_S(Q) \gets \topkargmax_{(s, a, v) \in \CN} \langle \Zbf(s \oplus a), \Zbf(Q) \rangle $.
            \State $\CN_V(Q) \gets \topkargmax_{(s, a, v) \in \CN} \langle \Zbf(v), \Zbf(Q) \rangle $.
            \State $\CN(Q) \gets \CN_S(Q) \cup \CN_V(Q)$.
            \State $\CC_{\CH}(Q) \gets \{\}$
            \While{$(|\CN(Q)| > 0) \lor (|\CC_{\CH}(Q)| < L)$}
                \State $\CC_{\CH}(Q) \gets \CC_{\CH}(Q) \cup  \argmax_{e \in \CE}{|\{n \in \CN(Q): n \in e\}|}$
            \EndWhile
            % \State $\CC_{\CH}(Q) \gets \argmin_{\CC \subseteq \CE} {| \CC |}$ such that $\CN(Q) \subseteq \bigcup_{e \in \CC} e$.
            % \State $\CC_{\CH}(Q) \gets \topargmax{L}_{e \subseteq \CC_{\CH}(Q)}{|e|}$.
            \State \Return{$\CC_{\CH}(Q)$}
        \EndProcedure
    \end{algorithmic}
\end{algorithm}

\subsubsection{Relevant Hyeredges as Context}
We form the relevant context as the set of hyperedges $\CC_{\CH}(Q) \subset \CE$ that minimally cover the relevant hypernodes, $\CN(Q) = \CN_S(Q) \cup \CN_V(Q)$. This is formulated as an optimization problem and solved in a greedy manner. Since the objective of minimizing the number of hyperedges is linear under a matroid constraint, the solution can be shown to be optimal~\citep{korte2011combinatorial}. Specifically, we maintain a dictionary that maps each hypernode $n \in \CN$ to the set of hyperedges that it is a part of, \ie, $\CE(n)$, where $e \in \CE(n) \implies n \in e$. In each iteration, we add the hyperedge that covers the largest number of uncovered nodes to the context and remove those nodes from further consideration. This process is repeated until either we have $L$ hyperedges or all the relevant nodes are covered. In this way, the context is constructed as a collection of up to $L$ hyperedges representing \textit{facts} relevant to the given query. By organizing information into hyperedges, OG-RAG is able to group related facts together, ensuring that the retrieved context is both compact and comprehensive, capturing all necessary facts to support accurate LLM responses, while optimizing for efficiency.

\subsubsection{Retrieval-Augmented Generation}

Finally, given a user query $Q$ and the relevant context as found above, we prompt the LLM $\CM$ to use this context to answer the query as $\CM(\CP(Q, \CC_{\CH}(Q))$, where $\CP$ denotes the corresponding textual prompt:

\noindent{\small\texttt{Given the context below, generate the answer to the given query. Note that the context is provided as a list of valid facts in a dictionary format.\\
Context: \textit{\textcolor{gray}{< Line-separated retrieved context $\CC_{\CH}(Q)$ >}} \\
Query: \textit{\textcolor{gray}{< User-defined query $Q$ >}}\\
Answer: 
}}

\subsection{Complexity Analysis}
Algorithm~\ref{alg:ograg} outlines the full procedure of the proposed method which consists of two main components: (1) \textsc{OG-Preprocess}, applied to the set of documents once, and (2) \textsc{OG-Retrieve}, used to retrieve the relevant context for each query.

\subsubsection{Query Complexity} Assume the context size for the LLM $\CM_0$ is $N_C$. The ontology $\CO$, which can be written in a JSON-LD or textual format, has a length $|\CO|$, where the attributes are mapped to their corresponding ranges in the natural language vocabulary. \textsc{OG-Preprocess} phase may involves several LLM calls depending on the number of document chunks, specifically, $(|\CD|+|\CO|)/N_C$ number of calls. We do not make any additional LLM calls during the querying time in the \textsc{OG-Retrieve} procedure.

\subsubsection{Time Complexity} We ignore the time taken by LLM calls while calculating the time complexity, as this is accounted for under query complexity. Thus, the time complexity of the \textsc{OG-Preprocess} step only involves the hypergraph transformation by flattening the mapped data. Let us assume we have $|\CI|$ factual-blocks derived from the documents, and each factual-block has a maximum length of $|F|_{max} = O(|\CO|)$. We consider two cases: \textbf{(1) Minimal or No Nesting}: In this case, the time complexity is determined by step 4 in the algorithm, leading to a complexity of $O(|\CO| |\CI|)$, \textbf{(2) Maximum Nesting}: In this scenario, step 4 may result in an empty set. Thus, each factual-block $F$ can be recursively flattened $\log |\CO|$ times while searching through the entire set, leading to a time complexity of $O(|\CI| |\CO| \log |\CO| )$.

\subsubsection{Space Complexity} The only storage required is for the hypergraph structure $\CH(\CI)$, which is directly proportional to the number of hyperedges $|\CE| = |\overline{\CI}|$. 

%% file: sec-setup.tex
\section{Experimental Setup}\label{sec:setup}
\textbf{Datasets.}
We evaluate OG-RAG across two distinct domain categories that involve specialized workflows: (a) Industrial workflows, with a focus on the agriculture domain, where precise, data-driven decisions are critical for crop management and resource allocation, and (b) Knowledge work, where we evaluate it on research and analysis tasks in the news domain. We avoid general domains like Wikipedia to mitigate potential data contamination in LLM training. For the agriculture domain, we utilize two proprietary high-quality datasets comprising of $85$ documents prepared by agriculture experts, focusing on the crop cultivation of Soybean and Wheat in India. For the news domain, we use the publicly available dataset from Multi-hop RAG~\citep{tang2024multihop},  filtered for 149 long-form articles (each over 2,000 words) focused on multi-faceted, complex news stories requiring detailed, contextually rich analysis. Please refer to Appendix~\ref{app:datasets} for exemplary excerpts from the datasets.

\textbf{Ontology.}
We use a semi-automated approach to construct the ontology for both domains, which reflects the broader applicability of OG-RAG in specialized workflows. For the agriculture domain, the ontology was generated using a proprietary ontology learning module, which was then reviewed and verified by multiple experts specializing in crop cultivation. For the news domain, we modify the existing Simple News and Press (SNaP) ontology\footnote{\url{https://iptc.org/thirdparty/snap-ontology/}}. Specifically, we simplify its structure by excluding certain classes, such as those related to images, videos, and the "stuff" hierarchy. Instead, we allow an asset to be linked to multiple events, and each event can be associated with multiple organizations and persons. The complete ontologies for both domains are provided in Appendix~\ref{app:ontology}.

\textbf{Large Language Models.}
We consider 4 large language models for zero-shot query answering while adding the retrieved context from different methods: 2 closed-box models~\footnote{\url{https://openai.com/index}\label{openai}} (GPT-4o-mini and GPT-4o) and 2 open-source models~\footnote{\url{https://ai.meta.com/blog/meta-llama-3-1/}} (Llama-3.1-8B and Llama-3.1-70B). These models have been chosen for their remarkable understanding and ability to reason in natural language. We consider $4096$ completion tokens and a temperature of $0$.

\textbf{Baselines.}
We compare OG-RAG against three leading retrieval-based methods, representing state-of-the-art approaches to context retrieval and generation, to demonstrate its effectiveness.

\begin{enumerate}[leftmargin=*]
    \item \textbf{RAG~\citep{lewis2020retrieval}:} RAG (Retrieval-Augmented Generation) retrieves query-relevant document chunks by embedding them into a vector space and then finding the context based on the maximum chunk-query similarity. 
    \item \textbf{RAPTOR~\citep{sarthi2024raptor}:} RAPTOR clusters document chunks into hierarchical structures and uses an LLM to summarize the clusters as additional context. For this experiment, we set the tree depth to $3$ and use the collapsed-tree retrieval strategy.
    \item \textbf{GraphRAG~\citep{edge2024graphrag}:} GraphRAG retrieves from a knowledge graph constructed using an LLM by extracting entities and relationships and clustering them into semantic communities. We use default graph construction prompts and local search with community level as $2$ for retrieval.
    % , retrieving query-relevant subgraphs based on entity similarity. 
    % \ks{fill} \hlc{I added more context please feel free to edit or add on this.}
\end{enumerate}

\noindent We use the text-embedding-3-small~\textsuperscript{\ref{openai}} as the sentence embedding function across all retrieval methods and GPT-4o as the LLM (\ie, $\CM_0$) for pre-processing. For each method, we find $\{2, 5\}$ similar contexts and select the one with the highest performance.

\textbf{Metrics}.
Building on the RAGAS framework~\citep{ragas},  we use the following metrics to assess the quality of the retrieved context and the generated responses while using text-embedding-3-small as the embedding model and GPT-4o as the LLM.
\begin{enumerate}[leftmargin=*]
    \item \textbf{Context Recall (C-Rec):} Proportion of claims in the ground-truth answer that can be attributed to the information present in the retrieved context.
    \item \textbf{Context Entity Recall (C-ERec):} Proportion of entities in the ground-truth answer that are present in the retrieved context.
    \item \textbf{Answer Similarity (A-Sim):} Similarity between the generated response and the ground-truth answer in the embedding space.
    \item \textbf{Answer Correctness (A-Corr):} A combination of answer similarity (defined above) and factual similarity, which is the F1-score between the claims in the ground-truth answer and those in the generated response.
    \item \textbf{Answer Relevance (A-Rel):} Measures how easily the original question can be inferred from the generated response. 
\end{enumerate}
\textbf{Ethics Consideration.} The survey was conducted in compliance with ACM's Publications Policy on Human Research. No personal data was collected, and all responses were anonymized. For the agriculture and news domain experiments, only publicly available or proprietary datasets were used, with no sensitive data involved, adhering to ethical research standards.

%% file: sec-results.tex
\section{Experiments}\label{sec:results}
\subsection{Query answering}

\subsubsection{Question Generation}

We generate a set of question/answer pairs using the RAGAS framework~\citep{ragas} to validate the factual accuracy of our proposed method. RAGAS prompts off-the-shelf LLM to generate questions of varying difficulty, each with the corresponding ground-truth answers and contexts. Specifically, we generate up to $100$ unique questions from RAGAS focused on multi-hop reasoning abilities, which is commonly required in specialized domain tasks. Examples of these generated questions, along with their ground-truth answers, are provided in Appendix~\ref{app:evalqs}.

\begin{table}[t]
    \centering
    \caption{Quality of contexts retrieved by different methods for domain-specific query-answering. We found the $95\%$ confidence interval to be $\le 0.05$ for all metrics, representing small margin of error. It is not reported here. The symbol `-' denotes that the computation did not complete within 1 day.}
    \label{tab:context}
    \resizebox{1.0\linewidth}{!}{%
    \begin{tabular}{c c cccccc cccccc cccccc}
        \toprule
        & \multirow{2}{*}{Method} & \multicolumn{5}{c}{Soybean} & \multicolumn{5}{c}{Wheat} & \multicolumn{5}{c}{News} \\
        \cmidrule(lr){3-7} \cmidrule(lr){8-12} \cmidrule(lr){13-17}
        % & & \multicolumn{3}{c}{Soybean} & \multicolumn{3}{c}{Wheat} \\
        & & C-Rec & C-ERec & A-Corr & A-Sim  & A-Rel & C-Rec & C-ERec & A-Corr & A-Sim  & A-Rel & C-Rec & C-ERec & A-Corr & A-Sim  & A-Rel \\
        \midrule 
        & RAG & 0.22 & 0.08 & 0.31 & 0.62 & 0.29 & 0.14 & 0.04 & 0.29 & 0.69 & 0.28 & 0.01 & 0.01 & 0.27 & 0.67 & 0.20\\
        & RAPTOR & 0.54 & 0.19 & 0.34 & 0.68 & 0.68 & 0.85 & 0.29 & 0.59 & 0.79 & 0.89 & \textbf{0.82} & 0.46 & 0.58 & 0.84 & 0.76 \\
        & GraphRAG & 0.41 & 0.14 & 0.26 & 0.63 & 0.63 & 0.78 & 0.05 & - & - & - & - & - & - \\
        & \textbf{OG-RAG} & \textbf{0.84} & \textbf{0.41} & \textbf{0.48} & \textbf{0.72} & \textbf{0.79} & \textbf{0.95} & \textbf{0.34} & 0.62 & 0.79 & 0.79 & \textbf{0.82} & \textbf{0.52} & 0.66 & 0.86 & 0.73\\
        \bottomrule
    \end{tabular}}
\end{table}

\subsubsection{Does OG-RAG help in retrieving useful contexts?}

A context is deemed useful for a query if it provides sufficient information to derive the ground-truth response. We evaluate this using Context Recall and Context Entity Recall. Table~\ref{tab:context} compares the performance of different retrieval methods across three datasets. OG-RAG outperforms the baselines in almost all cases, boosting the recall of correct claims by $55\%$ and recall of correct entities by $110\%$. The only exception is the News dataset where OG-RAG matches the context recall performance of RAPTOR but still delivers better performance.

\subsubsection{Does OG-RAG help generate factually accurate responses?}

A useful context should lead to more factual and precise response when incorporated into the query for various LLMs. We evaluate this by comparing how closely the generated responses/answers align with the ground-truth answer when added as context across different LLMs. Table~\ref{tab:answer} presents the results of response correctness, similarity, and relevance for the $3$ datasets. OG-RAG consistently outperforms the baselines, significantly improving answer correctness by $40\%$, and answer relevance by $16\%$. The only notable exception where OG-RAG slightly underperforms is in the Answer Relevance for Wheat and Soybean datasets in GPT-4o and Llama-3-70B. This is likely due to the broad scope of the retrieved context, which can sometimes introduce extraneous information. This can be possibly mitigated through further fine-tuning of the hypergraph retrieval mechanism, adjusting the level of detail to suit the complexity of the queries expected. We leave domain-specific optimization for future work, as the current approach already delivers good responses across all datasets.

\subsubsection{Is OG-RAG efficient?}

Finally, we demonstrate that OG-RAG is computationally efficient by comparing its pre-processing and per-query retrieval times with other methods across different datasets. Table~\ref{tab:time} shows that OG-RAG performs nearly as efficiently as a simple RAG method, with only a minimal increase of at most $2$ seconds during querying time despite being at least $100\%$ better in factual accuracy. OG-RAG is also shown to have significantly lower computational time than more competitive baselines such as RAPTOR and GraphRAG at both the pre-processing and query stages, particularly highlighted by a $~50\%$ drop in the pre-processing times. This efficiency is particularly critical for real-time applications, such as agricultural monitoring systems, legal research, and automated news fact-checking, where quick retrieval and processing of domain-specific knowledge is essential.

\begin{table}[t]
    \centering
    \caption{Quality of the answers generated by different LLMs using different retrieval methods. We found the $95\%$ confidence interval to be $\le 0.05$ for all metrics, so it is not reported here. The symbol `-' denotes that the computation did not complete within 1 day.}
    \label{tab:answer}
    \resizebox{1.0\linewidth}{!}{%
    \begin{tabular}{l c ccccc ccccc ccccc}
        \toprule
        & \multirow{2}{*}{Method} & \multicolumn{5}{c}{Soybean} & \multicolumn{5}{c}{Wheat} & \multicolumn{5}{c}{News} \\
        \cmidrule(lr){3-7} \cmidrule(lr){8-12} \cmidrule(lr){13-17}
        % & & \multicolumn{3}{c}{Soybean} & \multicolumn{3}{c}{Wheat} \\
        & & C-Rec & C-ERec & A-Corr & A-Sim  & A-Rel & C-Rec & C-ER & A-Corr & A-Sim  & A-Rel & C-Rec & C-ERec & A-Corr & A-Sim  & A-Rel \\
        \midrule 
        \multicolumn{2}{l}{\textit{Llama-3-8B}} \\
         & RAG & 0.22 & 0.07 & 0.26 & 0.59 & 0.22 & 0.14 & 0.05 & 0.26 & 0.65 & 0.23 & 0.01 & 0.01 & 0.15 & 0.52 & 0.08 \\
        & RAPTOR & 0.56 & 0.20 & 0.34 & \textbf{0.66} & 0.59 & 0.84 & \textbf{0.35} & \textbf{0.54} & \textbf{0.76} & 0.67 & 0.82 & 0.47 & \textbf{0.53} & 0.74 & 0.68 \\
        & GraphRAG & 0.46 & 0.11 &	0.26 & 	0.63 & 	0.52 &  &  & 0.43 & 0.35 & 0.27  & - & - & - & - & -\\
        & \textbf{OG-RAG} & \textbf{0.82} & \textbf{0.40} & \textbf{0.40} & 0.65 & \textbf{0.60} & \textbf{0.95} & 0.33 & \textbf{0.54} & 0.73 & \textbf{0.72} & 0.81 & 0.51 & 0.52 & \textbf{0.76} & \textbf{0.69} \\
        \midrule
        \multicolumn{2}{l}{\textit{Llama-3-70B}} \\
        & RAG & 0.24 & 0.06 & 0.27 & 0.59 & 0.19 & 0.14 & 0.03 & 0.26 & 0.65 & 0.14 & 0.01 & 0.01 & 0.17 & 0.58 & 0.09\\
        & RAPTOR & 0.55 & 0.23 & 0.41 & 0.70 & \textbf{0.64} & 0.85 & \textbf{0.39} & 0.58 & \textbf{0.77} & \textbf{0.75} & \textbf{0.82} & 0.47 & 0.39 & 0.72 & 0.64\\
        & GraphRAG &  &  &	0.30 & 	0.65 & 	0.55 &  &  & 0.47 & 0.37 & 0.29 & - & - & - & - & -\\
        & \textbf{OG-RAG} & \textbf{0.84} & \textbf{0.41} & \textbf{0.54} & \textbf{0.75} & 0.56 & \textbf{0.95} & 0.31 & \textbf{0.63} & \textbf{0.77} & 0.73 & 0.70 & \textbf{0.69} & \textbf{0.51} & \textbf{0.77} & \textbf{0.67} \\
        \midrule
        \multicolumn{2}{l}{\textit{GPT-4o-mini}} \\
        & RAG & 0.24 & 0.07 & 0.29 & 0.66 & 0.59 & 0.14 & 0.05 & 0.33 & 0.73 & 0.66 & 0.01 & 0.01 & 0.34 & 0.73 & 0.64\\
        & RAPTOR & 0.59 & 0.23 & 0.34 &	0.68 & \textbf{0.85} & 	0.84 & 0.36 &	0.51 &	0.77 &	\textbf{0.88} & 0.842995 &	0.364377 & 0.51 &	0.77 &	\textbf{0.88} \\
        & GraphRAG & 0.42 & 0.13 & 0.25 & 0.63 & 0.65 & 0.78 & 0.05 &	0.35 &	0.70 &	0.85 & - & - & - & - & - \\
        & \textbf{OG-RAG} & \textbf{0.83} & \textbf{0.41} & \textbf{0.48} & \textbf{0.72} & 0.77 & 0.95 & 0.33 & \textbf{0.62} & \textbf{0.78} & 0.85 & 0.81 & 	0.51 &	\textbf{0.62} & \textbf{0.78} & 0.85	\\
        \midrule
        \multicolumn{2}{l}{\textit{GPT-4o}} \\
        & RAG & 0.22 & 0.08 & 0.31 & 0.62 & 0.29 & 0.14 & 0.04 & 0.29 & 0.69 & 0.28 & 0.01 & 0.01 & 0.27 & 0.67 & 0.20\\
        & RAPTOR & 0.54 & 0.19 & 0.34 & 0.68 & 0.68 & 0.85 & 0.29 & 0.59 & \textbf{0.79} & \textbf{0.89} & 0.83 & 0.46 & 0.58 & 0.84 & \textbf{0.76} \\
        & GraphRAG & 0.41 & 0.14 & 0.26 & 0.63 & 0.63 & 0.82 & 0.52 &	0.35 &	0.70 &	0.86 & - & - & - & - & -  \\
        & \textbf{OG-RAG} & \textbf{0.84} & \textbf{0.41} & \textbf{0.48} & \textbf{0.72} & \textbf{0.79} & 0.95 & 0.34 & \textbf{0.62} & \textbf{0.79} & 0.79 & 0.82 & 0.52 & \textbf{0.66} & \textbf{0.86} & 0.73\\
        \bottomrule
    \end{tabular}}
\end{table}

\begin{table}[tb]
    \centering
    \caption{Efficiency of different retrieval methods on domain-specific query-answering. $T_{\text{pre}}$ and $T_{\text{query}}$ denote the average pre-processing and query time in seconds. We found the variance to be within $5$ seconds, so it is not reported here. }
    \label{tab:time}
    \resizebox{1.0\linewidth}{!}{%
    \begin{tabular}{l ccc ccc ccc }
        \toprule
        % Method & Soybean & Wheat & News \\
        \multirow{2}{*}{Method} & \multicolumn{3}{c}{Soybean} & \multicolumn{3}{c}{Wheat} & \multicolumn{3}{c}{News} \\
        \cmidrule(lr){2-4} \cmidrule(lr){5-7} \cmidrule(lr){8-10}
        & $T_{\text{pre}} \downarrow$ & $T_{\text{query}} \downarrow$ & $|\CC|$ & $T_{\text{pre}} \downarrow$ & $T_{\text{query}} \downarrow$ & $|\CC|$ & $T_{\text{pre}} \downarrow$ & $T_{\text{query}} \downarrow$ & $|\CC|$ \\
        \midrule 
        RAG & 11.41 & 2.49 & & 10.55  & 2.36 & & 449.21 & 3.56 \\
        RAPTOR & 71.66 & 4.81 & & 61.56 & 4.38 & & 1513.57 & 5.45 \\
        GraphRAG & 157.04 & 5.95 & & 307.37 & 5.65 & & >1 day & - \\
        \textbf{OG-RAG} & 29.61 & 3.75 & & 47.76 & 4.09 & & 655.15 & 4.12 \\
        % 10.27+19.33
        % 629.86
        \bottomrule
    \end{tabular}
    }
\end{table}

\subsection{Context attribution}

\subsubsection{Survey design}
To assess how effectively the proposed method aids humans in verifying facts within LLM-generated responses, we conduct a human study measuring the time taken to verify whether the given context supports the generated response. We randomly select $10$ queries from the agriculture dataset and present the responses generated by GPT-4o using both RAG and OG-RAG, each paired with their respective contexts. We exclude RAPTOR due to its content similarity with RAG, and GraphRAG due to its prohibitive context length. Participants are asked to evaluate the level of factual support the context provides for the response on a scale of 1-5. We also track the time each participant takes to complete this task. Each participant is shown $10$ questions, consisting of $5$ random queries, each paired with both RAG and OG-RAG responses and contexts in a randomized order. To ensure fairness, each query is presented an equal number of times across all participants. Examples of the survey design can be found in Appendix~\ref{app:survey}.

\begin{table}[t]
    \centering
    \caption{Comparison of the ease with which humans can attribute generated responses to the contexts produced by RAG and OG-RAG, presented with 95\% confidence intervals. 
    % \hlc{Make bar plots and also add Average Context Length with standard deviation}
    % \ks{maybe present as bar plots}
    }
    \label{tab:human}
    % \resizebox{0.9\linewidth}{!}{
    \begin{tabular}{c c c}
        \toprule
        Method & Time taken $\downarrow$ & Support [1-5] $\uparrow$ \\
        \midrule 
        RAG & $61.15 \pm 28.48$ & $2.67 \pm 0.30$ \\
        \textbf{OG-RAG} & \bm{$43.50 \pm 18.08$} & \bm{$3.46 \pm 0.19$} \\
        \bottomrule
    \end{tabular}
    % }
\end{table}

\subsubsection{Results}
A total of 16 participants, aged 18-34 and familiar with LLMs, took part in the survey. Table~\ref{tab:human} presents the average time taken and the level of support participants attributed to the contexts. We observed that OG-RAG significantly reduced the time required by $28.8\%$ and increased the human-attributed support by $29.6\%$ on average. These results demonstrate that OG-RAG not only enables faster fact verification but also provides more robust and clear contexts, making the system more user-friendly and reliable for context fact attribution.

% Table~\ref{tab:human} shows the results/ of the experiment. \ks{fill}. 
% \ks{
% \begin{enumerate}
%     \item Factuality on two domains: Agriculture and Retail with different metrics and baselines (RAG, GraphRAG, RAPTOR)
%     \item Human study on source attribution 
%     \item Rule deduction using self-made rules on Farm data
% \end{enumerate}
% }

\subsection{Factual Deduction}
\subsubsection{Deductive Facts}
We assess OG-RAG’s ability to enhance deductive reasoning in LLMs by evaluating how well it can generate new conclusions based on a set of predefined facts. These facts, grounded in domain-specific ontologies, provide the framework for reasoning tasks that require multi-step logic. Specifically, for this experiment we use six agricultural facts to deduce CO2 emissions, as this information is not directly available in the documents. These facts are partially derived from industry sources on the relationship between fossil fuels, pesticides, and greenhouse gases.~\footnote{Adapted in part from \url{https://www.panna.org/news/linking-fossil-fuels-and-pesticides-to-greenhouse-gases}}:
\begin{enumerate}[leftmargin=*]
    \item Farm area in the North Eastern Hill zone is 1 hectare or ha.
    \item Farm area in North Plain Hill zone is 2 hectares or ha.
    \item Herbicide production is calculated by multiplying the farm area by the recommended herbicide quantity.
    \item 1 kg of herbicide production results in 18.22–26.63 kg of CO2e emissions. 
    \item 1 kg of insecticide production results in 14.79–18.91 kg of CO2e emissions. 
    \item 1 kg of fungicide production results in 11.94–29.19 kg of CO2e emissions.
\end{enumerate}

\subsubsection{Question Generation}
To create the evaluation test set, we prompt GPT-4o following the RAGAS guidelines~\cite{ragas} to generate questions that require the application of deductive facts and a randomly sampled chunk from the ontology-mapped data to generate the responses. Specifically, we use the following prompt:
    
\noindent {\small\texttt{Given the following data and a set of deductive rules, generate a hard question that requires the application of the rules on the data to generate the answer. \\
    Data: \textcolor{gray}{< Domain-specific data >} \\
    Rules: \textcolor{gray}{< Fixed set of rules >} \\
    Question:
}}
    
Next, we make two additional LLM calls to generate the corresponding answer and assign a rating from 1 to 10, evaluating how well the question tests the application of the rules on the data to derive the answer. We select 10 questions that receive a rating of at least 7. A full list of generated questions is provided in Appendix~\ref{app:evalqs}. 

\subsubsection{Results}
Table~\ref{tab:rules} presents the results of factual deductions across two agriculture datasets, using GPT-4o and GPT-4o-mini as the underlying LLMs. In all cases, except two, the OG-RAG context substantially improves the correctness, similarity, and relevance of the generated answers compared to baseline methods. This demonstrates that OG-RAG is more effective at supporting deductive reasoning from a fixed set of facts. One exception is in the Soybean dataset for answer relevance which again points to a slightly less pertinent answer due to a broader retrieved context by OG-RAG. Overall, these results confirm that OG-RAG provides a more robust context for deducing new facts than alternative retrieval methods.

\begin{table}[t]
    \centering
    \caption{Comparison of different retrieval methods in their ability to support deductive reasoning from different LLMs.
    % \ks{experiment with some metrics}
    }
    \label{tab:rules}
    \resizebox{1.0\linewidth}{!}{%
    \begin{tabular}{l c ccc ccc}
        \toprule
        & \multirow{2}{*}{Method} & \multicolumn{3}{c}{Soybean} & \multicolumn{3}{c}{Wheat}\\
        \cmidrule(lr){3-5} \cmidrule(lr){6-8}
        % & & \multicolumn{3}{c}{Soybean} & \multicolumn{3}{c}{Wheat} \\
        & & A-Corr & A-Sim & A-Rel & A-Corr & A-Sim & A-Rel \\
        \midrule 
        % \multicolumn{2}{l}{\textit{Llama-3-8B}} \\
        %  & RAG \\
        % & RAPTOR \\
        % & GraphRAG \\
        % & \textbf{OG-RAG} \\
        % \midrule
        % \multicolumn{2}{l}{\textit{Llama-3-70B}} \\
        % & RAG \\
        % & RAPTOR \\
        % & GraphRAG \\
        % & \textbf{OG-RAG} \\
        % \midrule
        \multicolumn{2}{l}{\textit{GPT-4o-mini}} \\
        & RAG & 0.46 & 0.89 & 0.66 & 0.41 & 0.92 & 0.64 \\
        & RAPTOR & 0.42 & 0.89 & \textbf{0.81} & 0.50 & 0.92 & 0.74\\
        & GraphRAG & 0.44 &	0.91 &	0.83 & 0.49 & 0.93 & 0.82 \\
        & \textbf{OG-RAG} & \textbf{0.50} & \textbf{0.92} & 0.75 & \textbf{0.53} & \textbf{0.94} & \textbf{0.83} \\
        \midrule
        \multicolumn{2}{l}{\textit{GPT-4o}} \\
        & RAG & 0.44 & 0.90 & 0.56 & 0.42 & 0.92 & 0.54 \\
        & RAPTOR & 0.01 & 0.11 & 0.03 & 0.41 & 0.91 & 0.74 \\
        & GraphRAG & 0.48 & 0.92 &	\textbf{0.84} & 0.44 & 0.90 & 0.73 \\
        & \textbf{OG-RAG} & \textbf{0.56} & \textbf{0.92} & 0.75 & \textbf{0.47} &	\textbf{0.94} &	\textbf{0.83} \\
        \bottomrule
    \end{tabular}
    }
\end{table}

% \subsection{Hyperparameter analysis}
% We also analyze the effect of hyperparameters on the performance of our method by changing the number of selected hypernodes $k$ and the number of selected hyperedges $L$. Table~\ref{tab:hyperparams} show the change in the 

%% file: sec-conclusion.tex
\section{Conclusion}~\label{sec:conclusion}
In this work, we study the problem of domain adaptation of LLMs using ontology-grounded retrieval-augmented generation. We introduce OG-RAG, a novel hypergraph-based retrieval method that retrieves query-relevant context from documents by structuring their facts as a hypergraph using a domain-specific ontology. OG-RAG has wide applicability in domains which include industrial workflows in healthcare, legal, and agricultural sectors, among others as well as knowledge-driven tasks like news journalism, investigative research, consulting, and more. Through extensive experiments on two agriculture datasets and a news dataset, we demonstrate that OG-RAG significantly improves the factual accuracy of LLM-generated responses, while also enabling faster attribution of answers to their supporting context and more effectively deducing conclusions from domain facts.  We recommend that LLMs have better ways to incorporate controlled vocabulary and structured evidence retrieval through fixed ontologies, as this not only enhances user comprehension of generated responses but also facilitates smoother integration of LLMs into industrial workflows and knowledge work. By offering greater flexibility and control over how context is retrieved and utilized, OG-RAG paves the way for more adaptable and reliable language systems. For future work we encourage to explore automated or semi-automated ontology construction techniques to build these frameworks in an end-to-end fashion, ensuring broader applicability of retrieval-augmented models across diverse domains. 

%% file: sec-appendix.tex
\appendix
\section*{Appendix}

\section{Dataset Examples}\label{app:datasets}
\subsection{Soybean}
\ovalbox{
\parbox{0.8\linewidth}{
    ---
    title: SOYBEAN: AN INTRODUCTION Classification of States in six major Soybean growing zones: Soybean is majorly grown in the following areas
    ---
    \\
    \# SOYBEAN: AN INTRODUCTION Classification of States in six major Soybean growing zones: Soybean is majorly grown in the following areas
    \\
    \#\# TILLAGE
    \\
    • Deep ploughing is essential during summer, after harvesting the Rabi crop. This facilitates exposing the hibernating insects to extreme heat and predatory birds as well as movement of nutrients and infiltration of soil water. Therefore, one deep ploughing once in 3-4 years, otherwise one normal ploughing in summer followed by 2 criss-cross harrowing or cultivation for breaking of soil clods will make ideal seed bed for a good crop of soybean cultivation is recommended. Also, sub-soiling operation once in 4-5 years at an interval of 10 meter, break the compactness of the sub-soil and also facilitate infiltration of rainwater which is useful for un-interrupted crop growth even during drought period also.
}
}

\subsection{Wheat}
\ovalbox{
\parbox{0.8\linewidth}{
    \#\# CLIMATE
    \\
    Wheat crop has wide adaptability. It is primarily a temperate crop but is widely cultivated in subtropical regions and is grown even in some tropical countries.
    \\
    • Ideal temperature for germination of wheat seeds is around 20-23 o C though these can germinate in temperature range of 3.5 to 35 o C.
    \\
    • During the heading and flowering stages, excessively high or low temperatures and drought are harmful to wheat.
    \\
    • The temperature conditions at the time of grain filling and development are very crucial for yield.
    \\
    • Temperatures above 25 o C during grain filling and development period tend to depress grain weight. When temperatures are high, too much energy is lost through the process of transpiration by the plants and the reduced residual energy results in poorer grain formation and lower yields.
}
}

\subsection{News}
\ovalbox{
\parbox{0.8\linewidth}{
    \# Raiders vs. Lions live score, updates, highlights from NFL 'Monday Night Football' game\\
    **author**: Dan Treacy\\
    **source**: Sporting News\\
    **published\_at**: 2023-10-30T22:20:03+00:00\\
    **category**: sports\\
    **url**: https://www.sportingnews.com/us/nfl/news/
    raiders-lions-live-score-highlights-monday-night-football/d022b1d62b18af8a70c516f4
    \\
    The Lions just needed to get themselves back in the win column after a blowout loss in Baltimore, and they did just that in front of their home fans on Monday night.
    \\
    Detroit rolled to a 26-14 victory over the Raiders in a game that felt much more one-sided than the score indicates.
    \\
    The Lions thoroughly outplayed the Raiders, out-gaining Las Vegas by 329 yards, but critical mistakes by Detroit left points on the board. The offense struggled to turn red zone opportunities into touchdowns in the first half, and two turnovers deep in Raiders territory -- including a pick-six by Jared Goff -- kept Las Vegas in the game in the second half.
    MORE: What to know about Lions' retro 'Honolulu blue' helmets
}
}

\section{Ontology}

\subsection{Ontology Prompt}\label{app:ontology_prompt}

\texttt{  
Here is a context definition for wheat crop cultivation ontology.
\\
Context Definition: 
\\
\{context\_definition\}
\\
-----------------
\\
Generate a JSON-LD using the following data and the above context definition for crop cultivation ontology.\\
Use '@graph' object namespace for the data in JSON-LD.\\
Be comprehensive and make sure to fill all of the data. \\
Keep nesting to the minimum and still be able to disambiguate.\\
If there are multiple subfields enumerated in a 'List' namespace then do not combine them in a single subfield, keep them as separate subfields to disambiguate.\\
Ensure that you populate all items in the 'List' namespace, do not leave any item.\\
Do not include any explanations or apologies in your response.\\
Do not add any other text other than the generated JSON-LD in your response.\\
Generate in Json format.\\
----------------------\\
Data:\\
\\
\{data\}\\
---------------------\\
JSON-LD json:
}

\subsection{Examples}\label{app:ontology}

\begin{figure*}[!ht]
    \centering
    \input{news-ontology}
    \caption{News ontology}
    \label{fig:news_ontology}
\end{figure*}
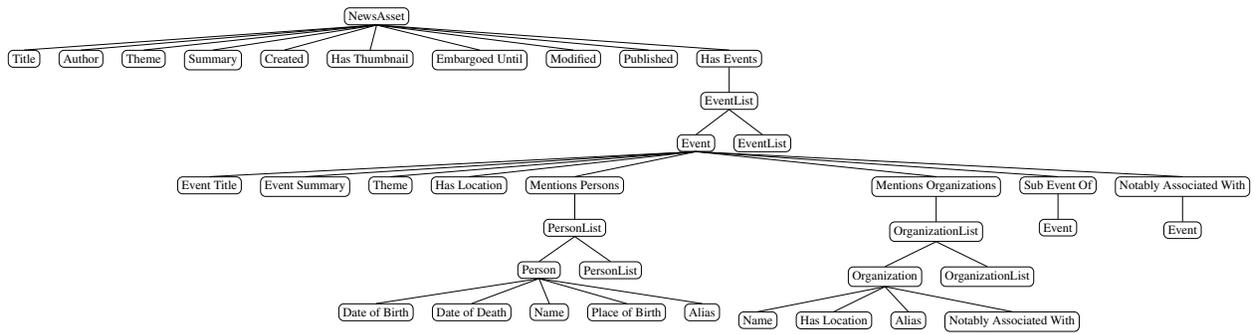

Figures~\ref{fig:agri_ontology} and ~\ref{fig:news_ontology} show agriculture and news ontologies used in the work. 

\section{Evaluation Questions}\label{app:evalqs}

\subsection{Soybean}
\begin{QAblock}
\textbf{Question:} Which pest can be controlled with Imidacloprid 48 FS pesticide in soybean crops? \\
\textbf{OG-RAG Answer:} The pest that can be controlled with Imidacloprid 48 FS pesticide in soybean crops is the White Grub.
\end{QAblock}

\begin{QAblock}
\textbf{Question:} What type of soil should be avoided for growing soybeans due to its medium to high nutrient holding capacity and medium water holding capacity?
\\
\textbf{RAPTOR Answer:} The context does not explicitly mention a specific type of soil that should be avoided for growing soybeans due to its medium to high nutrient holding capacity and medium water holding capacity. However, it does provide recommendations for managing different soil types:

- In saline soils (pH >7.5), gypsum at 150-200 kg/ha should be applied along with organic manure.
- In acidic soils (pH <6.5), lime at 600 kg/ha is recommended.

Based on this information, it can be inferred that while soybeans can be grown in various soil types, specific amendments are needed for saline and acidic soils to optimize growth. Therefore, no particular soil type is explicitly advised to be avoided, but appropriate soil management practices are necessary for different soil conditions.
\\
\textbf{OG-RAG Answer:} Sandy soil should be avoided for growing soybeans due to its medium to high nutrient holding capacity and medium water holding capacity.
\end{QAblock}

\begin{QAblock}
\textbf{Question:} What is the name of the crop that requires seeds with a germination rate of at least 80\%, should be harvested when 85-90\% pods have turned brown and lost their green color, and should be stored in a cool, dry place using hermetic storage bags for long-term storage?
\\
\textbf{RAPTOR Answer:} The context provided does not mention a crop that requires seeds with a germination rate of at least 80\%, should be harvested when 85-90\% pods have turned brown and lost their green color, and should be stored in a cool, dry place using hermetic storage bags for long-term storage. The context specifically discusses soybean harvesting, threshing, and storage practices, including a germination rate of at least 70\%, harvesting when 95\% of pods change color, and storage recommendations. Therefore, based on the given context, there is no crop that matches the criteria described in the question.
\\
\textbf{OG-RAG Answer:} The name of the crop is Soybean.
\end{QAblock}

\begin{QAblock}
\textbf{Question:} Which soybean crop is recommended for the Central Zone and has an early sowing time on January 1st, 2016?
\\
\textbf{RAPTOR Answer:} The context provided does not mention any soybean crop recommended for the Central Zone with an early sowing time on January 1st, 2016. The recommended sowing times for the Central Zone are from 20th June to 5th July. Therefore, there is no soybean crop recommended for the Central Zone with a sowing time of January 1st, 2016.
\\
\textbf{OG-RAG Answer:} The soybean crop recommended for the Central Zone with an early sowing time starting on January 1st, 2016, is the variety "JS 20-69."
\end{QAblock}

% \begin{QAblock}
% \textbf{Question:} What is the minimum germination rate for soybean seeds to meet the requirements for soybean crops?
% \\
% \textbf{OG-RAG Answer:} The minimum germination rate for soybean seeds to meet the requirements for soybean crops is at least 80\%.
% \end{QAblock}

% \begin{QAblock}
% \textbf{Question:} What are the requirements for soybean growth in Madhya Pradesh, including seed germination rate, harvesting and storage guidelines, and the growing zone?
% \\
% \textbf{OG-RAG Answer:} For soybean growth in Madhya Pradesh, the requirements are as follows:\n\n- **Seed Germination Rate**: Seeds should have a germination rate of at least 80\%.\n\n- **Harvesting Guidelines**: Harvest when 85-90\% of the pods have turned brown and lost their green color. The moisture content of the seeds should be around 10-12\%.\n\n- **Storage Guidelines**: Store in a cool, dry place. Use hermetic storage bags for long-term storage.\n\n- **Growing Zone**: Madhya Pradesh.\end{QAblock}

% \begin{QAblock}
% \textbf{Question:} What is the minimum percentage of sprouted seeds needed to meet the germination rate requirement for soybeans?
% \\
% \textbf{Answer:} The minimum percentage of sprouted seeds needed to meet the germination rate requirement for soybeans is 70\%.
% \end{QAblock}

\begin{QAblock}
\textbf{Question:} What is the temperature range for soybean germination and how does it compare to the temperature requirements for flowering and grain filling?
\\
\textbf{RAPTOR Answer:} The provided context does not specify the exact temperature range required for soybean germination, flowering, or grain filling. Therefore, based on general agronomic knowledge:

1. **Soybean Germination**: Soybeans typically germinate best at soil temperatures between 15°C to 30°C (59°F to 86°F). Optimal germination usually occurs around 25°C (77°F).

2. **Flowering and Grain Filling**: For flowering and grain filling, soybeans generally require slightly higher temperatures. The optimal temperature range for these stages is typically between 20°C to 30°C (68°F to 86°F). Temperatures above 30°C (86°F) can negatively impact these processes, especially if they are sustained over long periods.

In summary, while soybean germination can occur at a range of temperatures starting from around 15°C, the optimal conditions for flowering and grain filling are generally within a similar but slightly higher range, emphasizing the importance of maintaining moderate temperatures throughout the growing season for optimal yield.
\\
\textbf{OG-RAG Answer:} The temperature range for soybean germination is 15-30°C. For flowering, the temperature requirement is 20-30°C, and for grain filling, it is also 20-30°C. This means that the temperature range for germination is slightly broader on the lower end compared to flowering and grain filling, which have the same temperature requirements.
\end{QAblock}

\subsection{Wheat}

% \begin{QAblock}
% \textbf{Question:} Which regions are included in the Northern Hills Zone for wheat cultivation in Jammu and Kashmir?
% \\\               \textbf{OG-RAG Answer:} The Northern Hills Zone for wheat cultivation includes the region of Jammu and Kashmir.
% \end{QAblock}

\begin{QAblock}
\textbf{Question:} What is the recommended moisture level for harvesting wheat and how should the grain be dried before storage?
\\\               \textbf{OG-RAG Answer:} The recommended moisture level for harvesting wheat is 25\%. Before storage, the grain should be dried by spreading it on tarpaulin plastic sheets in bright sunlight to a moisture level below 12%.
\end{QAblock}

% \begin{QAblock}
% \textbf{Question:} What are the recommended practices for wheat grain harvesting and drying before storage?
% \\\               \textbf{OG-RAG Answer:} The recommended practices for wheat grain harvesting and drying before storage are as follows:

% 1. **Harvesting:**
%    - Harvest when the moisture level of the grain is 25%.
%    - Manual harvesting is common, but a combine harvester can be used for quick harvesting to avoid losses due to shattering and lodging.
%    - Harvest wheat 4-5 days before it is dead ripe.
%    - Morning is the best time for harvesting.
%    - For manual harvesting, make bundles and dry them for 3-4 days before threshing with a thresher.

% 2. **Drying:**
%    - Before storage, dry the grain by spreading it on tarpaulin plastic sheets in bright sunlight.
%    - Ensure the moisture level is reduced to below 12% before storage.
% \end{QAblock}

\begin{QAblock}
\textbf{Question:} Why conduct a seed germination test for wheat in relation to food production?
\\\               \textbf{OG-RAG Answer:} Conducting a seed germination test for wheat is essential to ensure a minimum germination rate of 85\%. This is crucial for achieving an optimum plant population, which directly impacts the yield. By verifying the germination status of seeds, farmers can ensure that they plant seeds that are likely to grow successfully, leading to better food production and higher yields.
\end{QAblock}

\begin{QAblock}
\textbf{Question:} What are the recommended practices for harvesting and drying wheat grains in the NEP Zone before storage?
\\\               \textbf{OG-RAG Answer:} The recommended practices for harvesting and drying wheat grains in the North Western Plains (NEP) Zone before storage are as follows:

1. **Harvesting:**
   - Harvest when the moisture level of the grain is 25%.
   - Wheat should be harvested 4-5 days before it is dead ripe.
   - Morning is the best time for harvesting.
   - For quick harvesting, use a combine harvester to avoid losses due to shattering and lodging.
   - In case of manual harvesting, make bundles and dry them for 3-4 days before threshing with a thresher.

2. **Drying:**
   - Before storage, dry the grain by spreading it on tarpaulin plastic sheets in bright sunlight.
   - Ensure the moisture level is reduced to below 12\% before storage.
\end{QAblock}

% \begin{QAblock}
% \textbf{Question:} What is the minimum germination percentage required for a wheat seed germination test in the NEP Zone?
% \\\               \textbf{OG-RAG Answer:} The minimum germination percentage required for a wheat seed germination test, as mentioned in the context, is 85\%. This requirement is not specific to a particular zone, such as the NEP Zone, but is a general guideline for wheat.
% \end{QAblock}

% \begin{QAblock}
% \textbf{Question:} What are the recommended moisture level and time of day for manual wheat crop harvesting?
% \\\               \textbf{OG-RAG Answer:} The recommended moisture level for manual wheat crop harvesting is 25\%. The best time of day for harvesting is in the morning.
% \end{QAblock}

% \begin{QAblock}
% \textbf{Question:} What are the recommended moisture level and best time for manual wheat harvesting?
% \\\               \textbf{OG-RAG Answer:} The recommended moisture level for manual wheat harvesting is 25\%. The best time for manual harvesting is in the morning.
% \end{QAblock}

% \begin{QAblock}
% \textbf{Question:} Which hilly areas in Tamil Nadu and Kerala are included in the Southern Hills Zone for wheat cultivation?
% \\\               \textbf{OG-RAG Answer:} The hilly areas in Tamil Nadu and Kerala are included in the Southern Hills Zone for wheat cultivation.
% \end{QAblock}

\begin{QAblock}
\textbf{Question:} What are the recommended practices for harvesting and drying wheat grains in the Northern Hills Zone before storage?
\\                
\textbf{OG-RAG Answer:} In the Northern Hills Zone, the recommended practices for harvesting and drying wheat grains before storage are as follows:

1. **Harvesting:**
   - Harvest when the moisture level of the grain is 25%.
   - Wheat is generally manually harvested, but a combine harvester can be used for quick harvesting to avoid losses due to shattering and lodging.
   - Harvest 4-5 days before the wheat is dead ripe.
   - Morning is the best time for harvesting.
   - In case of manual harvesting, make bundles and dry them for 3-4 days before threshing with a thresher.

2. **Drying:**
   - Before storage, dry the grain by spreading it on tarpaulin plastic sheets in bright sunlight to reduce the moisture level below 12\%.
\end{QAblock}

\subsection{News}

               \begin{QAblock}
                \textbf{Question:} Which earbuds are stylish and see-through?
                \\               \textbf{OG-RAG Answer:} The earbuds that are stylish and see-through are the "Nothing Ear Stick."
                \end{QAblock}

               \begin{QAblock}
                \textbf{Question:} What else debuts with 'Hurricane Season' on Netflix Nov. 1?
                \\           
                \textbf{OG-RAG Answer:} On Netflix Nov. 1, along with "Hurricane Season," the following titles debut:

- Locked In (Netflix Film)
- Mysteries of the Faith (Netflix Documentary)
- Nuovo Olimpo (Netflix Film)
- Till Murder Do Us Part: Soering vs. Haysom (Netflix Documentary)
- Wingwomen (Netflix Film)
- 13 Going on 30
- 13 Hours: The Secret Soldiers of Benghazi
- 60 Days In: Season 5
- The Addams Family
- The Big Lebowski
- The Change-Up
- Desperado
- Downsizing
- Drag Me to Hell
- Love in the Wild: Season 1
- Madea’s Family Reunion
- The Mummy (2017)
- Paul Blart: Mall Cop
- Paul Blart: Mall Cop 2
- Pitch Perfect
- The Scorpion King
- Scott Pilgrim vs. the World
- Sherlock Gnomes
- Six Feet Under: Seasons 1-5
- Sixteen Candles
- The Social Network
- Ted 2
- Victorious: Season 3
- Whiplash
                \end{QAblock}

               \begin{QAblock}
                \textbf{Question:} How do price-match policies curb customer loss in big sales?
                \\               \textbf{OG-RAG Answer:} Price-match policies help curb customer loss during big sales by ensuring that customers can get the best price without needing to shop around at multiple retailers. By offering to match lower prices found elsewhere, retailers like Best Buy, Walmart, Dell, eBay, and Lowe’s aim to keep customers satisfied and loyal, preventing them from purchasing from competitors. This strategy is particularly effective during major sales events like Black Friday and Cyber Monday, when price competition is intense. By guaranteeing competitive pricing, retailers can maintain their customer base and potentially attract new customers who are looking for the best deals.
                \end{QAblock}

               \begin{QAblock}
                \textbf{Question:} Who wrote about AI risks and the creative class?
                \\               \textbf{OG-RAG Answer:} Daniel Tencer wrote about AI risks and the creative class.
                \end{QAblock}

               % \begin{QAblock}
               %  \textbf{Question:} Which industry had 9,000+ layoffs tied to Epic Games in 2023?
               %  \\               \textbf{OG-RAG Answer:} The industry with 9,000+ layoffs in 2023 is the video games industry. However, the layoffs are not specifically tied to Epic Games. The context mentions "Over 9,000 people in video games were laid off in 2023," but it does not directly associate these layoffs with Epic Games.
               %  \end{QAblock}

               \begin{QAblock}
                \textbf{Question:} Who hit the Hawks' game-winning buzzer-beater vs. the Taipans?
                \\               \textbf{OG-RAG Answer:} Tyler Harvey hit the Hawks' game-winning buzzer-beater vs. the Taipans.
                \end{QAblock}

               \begin{QAblock}
                \textbf{Question:} Which Chrome extension by Steven Tey aids in sharing AI replies?
                \\               \textbf{OG-RAG Answer:} The Chrome extension by Steven Tey that aids in sharing AI replies is called "ShareGPT."
                \end{QAblock}

               % \begin{QAblock}
               %  \textbf{Question:} Which 2023 event offers Apple deals?
               %  \\               \textbf{OG-RAG Answer:} The context provided does not mention any event offering Apple deals in 2023.
               %  \end{QAblock}

               \begin{QAblock}
                \textbf{Question:} Which artists thrived under Michelle Jubelirer at Capitol Music Group?
                \\               \textbf{OG-RAG Answer:} The artists that thrived under Michelle Jubelirer at Capitol Music Group include Ice Spice and The Beatles.
                \end{QAblock}

               \begin{QAblock}
                \textbf{Question:} Why are JLab earphones seen as a budget-friendly backup with EQ presets?
                \\               \textbf{OG-RAG Answer:} JLab earphones are considered a budget-friendly backup because they offer great value by including three different EQ presets directly into the earphones. This feature enhances the listening experience without a high cost, making them an attractive option for those seeking an affordable alternative to their main wireless earphones.
                \end{QAblock}

\subsection{Factual Deduction}

\begin{QAblock}
                \textbf{Question:} Calculate the total CO2e emissions from herbicide production for a farm located in the North Eastern Plains Zone, using the recommended herbicide for Phalaris minor. Assume the farm area is 1 hectare. Provide the range of CO2e emissions based on the given data and rules.
                \\                \textbf{OG-RAG Answer:} To calculate the total CO2e emissions from herbicide production for a farm located in the North Eastern Plains Zone, we need to follow these steps:

1. **Identify the Recommended Herbicide and Quantity:**
   - The recommended herbicide for Phalaris minor is Clodinafop.
   - The recommended quantity is 60 gm a.i. (active ingredient) per hectare.

2. **Calculate the Total Herbicide Required:**
   - Since the farm area is 1 hectare, the total herbicide required is 60 grams.

3. **Convert Grams to Kilograms:**
   - 60 grams is equivalent to 0.06 kilograms.

4. **Calculate CO2e Emissions:**
   - The CO2e emissions for herbicide production range from 18.22 to 26.63 kg of CO2e per kilogram of herbicide.
   - For 0.06 kg of herbicide, the CO2e emissions are calculated as follows:
     - Lower bound: 0.06 kg * 18.22 kg CO2e/kg = 1.0932 kg CO2e
     - Upper bound: 0.06 kg * 26.63 kg CO2e/kg = 1.5978 kg CO2e

Therefore, the total CO2e emissions from herbicide production for the farm are in the range of 1.0932 to 1.5978 kg CO2e.
                \end{QAblock}

\begin{QAblock}
\textbf{Question:} Calculate the total potential CO2e emissions from herbicide, insecticide, and fungicide production for a farm located in the North Plain zone, assuming the recommended herbicide quantity is 2 kg per hectare, insecticide quantity is 1.5 kg per hectare, and fungicide quantity is 1 kg per hectare. Use the average CO2e values for your calculations.
                \\               
\textbf{OG-RAG Answer:} To calculate the total potential CO2e emissions from herbicide, insecticide, and fungicide production for a farm located in the North Plain zone, we need to follow these steps:

1. **Determine the farm area**: The farm area in the North Plain Hill zone is 2 hectares.

2. **Calculate the total quantity of each type of chemical used**:
   - Herbicide: 2 kg/ha × 2 ha = 4 kg
   - Insecticide: 1.5 kg/ha × 2 ha = 3 kg
   - Fungicide: 1 kg/ha × 2 ha = 2 kg

3. **Use the average CO2e values for each type of chemical**:
   - Herbicide: Average CO2e = (18.22 + 26.63) / 2 = 22.425 kg CO2e/kg
   - Insecticide: Average CO2e = (14.79 + 18.91) / 2 = 16.85 kg CO2e/kg
   - Fungicide: Average CO2e = (11.94 + 29.19) / 2 = 20.565 kg CO2e/kg

4. **Calculate the total CO2e emissions for each type of chemical**:
   - Herbicide: 4 kg × 22.425 kg CO2e/kg = 89.7 kg CO2e
   - Insecticide: 3 kg × 16.85 kg CO2e/kg = 50.55 kg CO2e
   - Fungicide: 2 kg × 20.565 kg CO2e/kg = 41.13 kg CO2e

5. **Calculate the total potential CO2e emissions**:
   - Total CO2e = 89.7 kg + 50.55 kg + 41.13 kg = 181.38 kg CO2e

Therefore, the total potential CO2e emissions from herbicide, insecticide, and fungicide production for the farm in the North Plain zone is 181.38 kg CO2e.
\end{QAblock}

\begin{QAblock}
\textbf{Question:} Calculate the total CO2e emissions from herbicide production for a farm in the North Eastern Hill zone, assuming the recommended herbicide quantity is 3 kg per hectare. Use the range provided for CO2e emissions per kg of herbicide to determine the minimum and maximum possible emissions.
\\               
\textbf{OG-RAG Answer:} To calculate the total CO2e emissions from herbicide production for a farm in the North Eastern Hill zone, we need to follow these steps:

1. **Determine the farm area**: According to the context, the farm area in the North Eastern Hill zone is 1 hectare.

2. **Calculate the total herbicide quantity**: The recommended herbicide quantity is 3 kg per hectare. Therefore, for a 1-hectare farm, the total herbicide quantity is:
   \[
   3 \, \text{kg/ha} \times 1 \, \text{ha} = 3 \, \text{kg}
   \]

3. **Calculate the CO2e emissions**: The CO2e emissions per kg of herbicide range from 18.22 to 26.63 kg of CO2e. Therefore, the total CO2e emissions for 3 kg of herbicide are:

   - **Minimum emissions**:
     \[
     3 \, \text{kg} \times 18.22 \, \text{kg CO2e/kg} = 54.66 \, \text{kg CO2e}
     \]

   - **Maximum emissions**:
     \[
     3 \, \text{kg} \times 26.63 \, \text{kg CO2e/kg} = 79.89 \, \text{kg CO2e}
     \]

Therefore, the total CO2e emissions from herbicide production for the farm in the North Eastern Hill zone range from 54.66 kg to 79.89 kg of CO2e.
\end{QAblock}

% \begin{QAblock}
% \textbf{Question:} Calculate the total CO2e emissions from herbicide production for a farm located in the North Eastern Plains Zone, assuming the recommended herbicide quantity is 3 kg per hectare. Use the range provided for CO2e emissions per kg of herbicide.
%                 \\
% \textbf{OG-RAG Answer:} To calculate the total CO2e emissions from herbicide production for a farm located in the North Eastern Plains Zone, we need to follow these steps:

% 1. **Determine the Farm Area**: The context does not provide a specific farm area for the North Eastern Plains Zone. However, we can assume a hypothetical farm area for calculation purposes. Let's assume a farm area of 1 hectare for simplicity.

% 2. **Calculate the Total Herbicide Quantity**: 
%    - Recommended herbicide quantity = 3 kg per hectare
%    - Total herbicide quantity for 1 hectare = 3 kg

% 3. **Calculate CO2e Emissions**:
%    - CO2e emissions per kg of herbicide range from 18.22 to 26.63 kg of CO2e.
%    - For 3 kg of herbicide, the total CO2e emissions would be:
%      - Minimum emissions = 3 kg * 18.22 kg CO2e/kg = 54.66 kg CO2e
%      - Maximum emissions = 3 kg * 26.63 kg CO2e/kg = 79.89 kg CO2e

% Therefore, the total CO2e emissions from herbicide production for a 1-hectare farm in the North Eastern Plains Zone would range from 54.66 kg to 79.89 kg of CO2e.
% \end{QAblock}

\section{Survey design}\label{app:survey}
\begin{figure*}[]
    \centering
    \hspace*{\fill}
    \subfloat[Introduction page]{\includegraphics[width=0.3\linewidth, trim={10cm 18cm 10cm 8cm},clip]{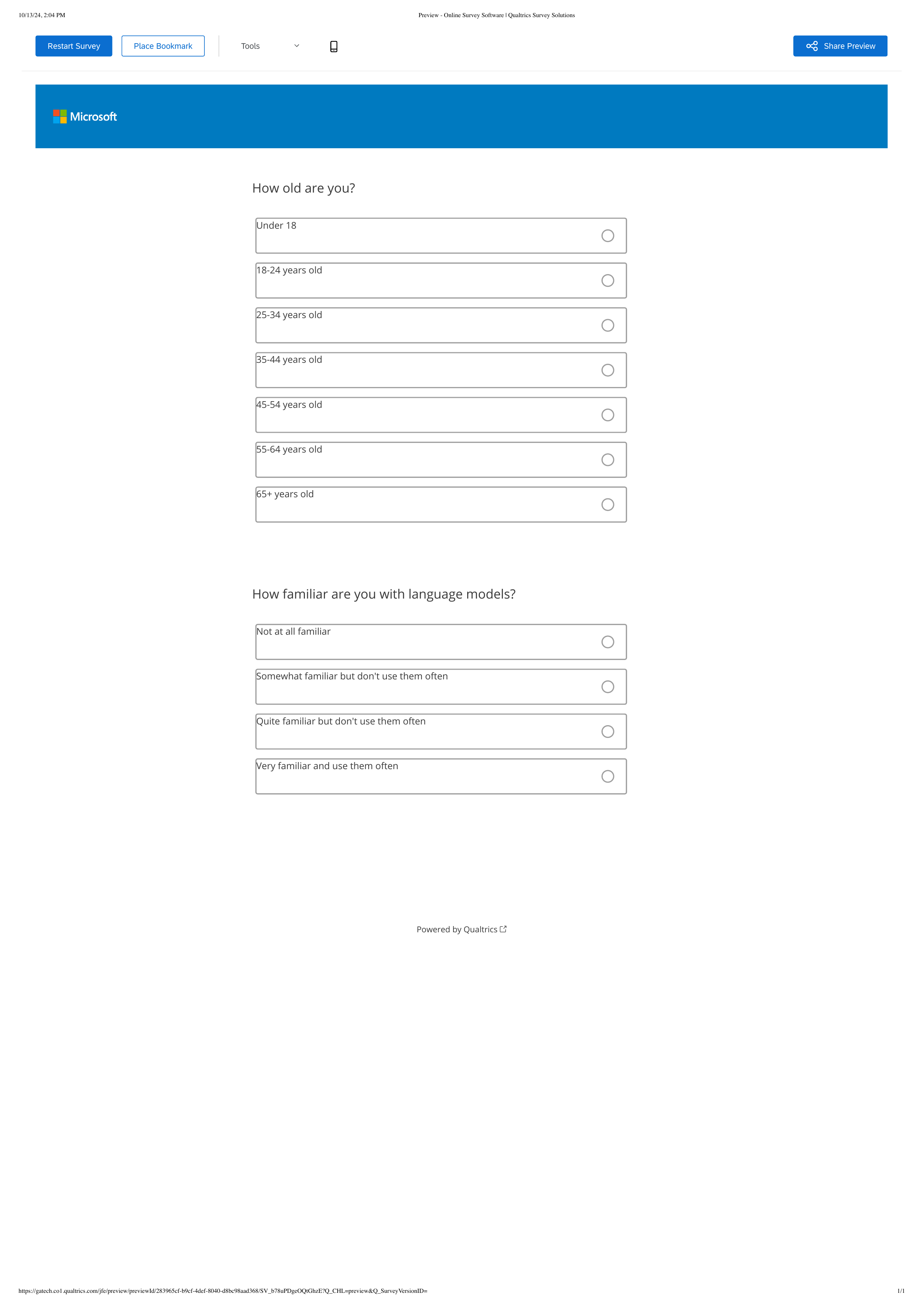}}
    \hfill
    \subfloat[RAG context attribution]{\includegraphics[width=0.3\linewidth, trim={10cm 10cm 10cm 8cm},clip]{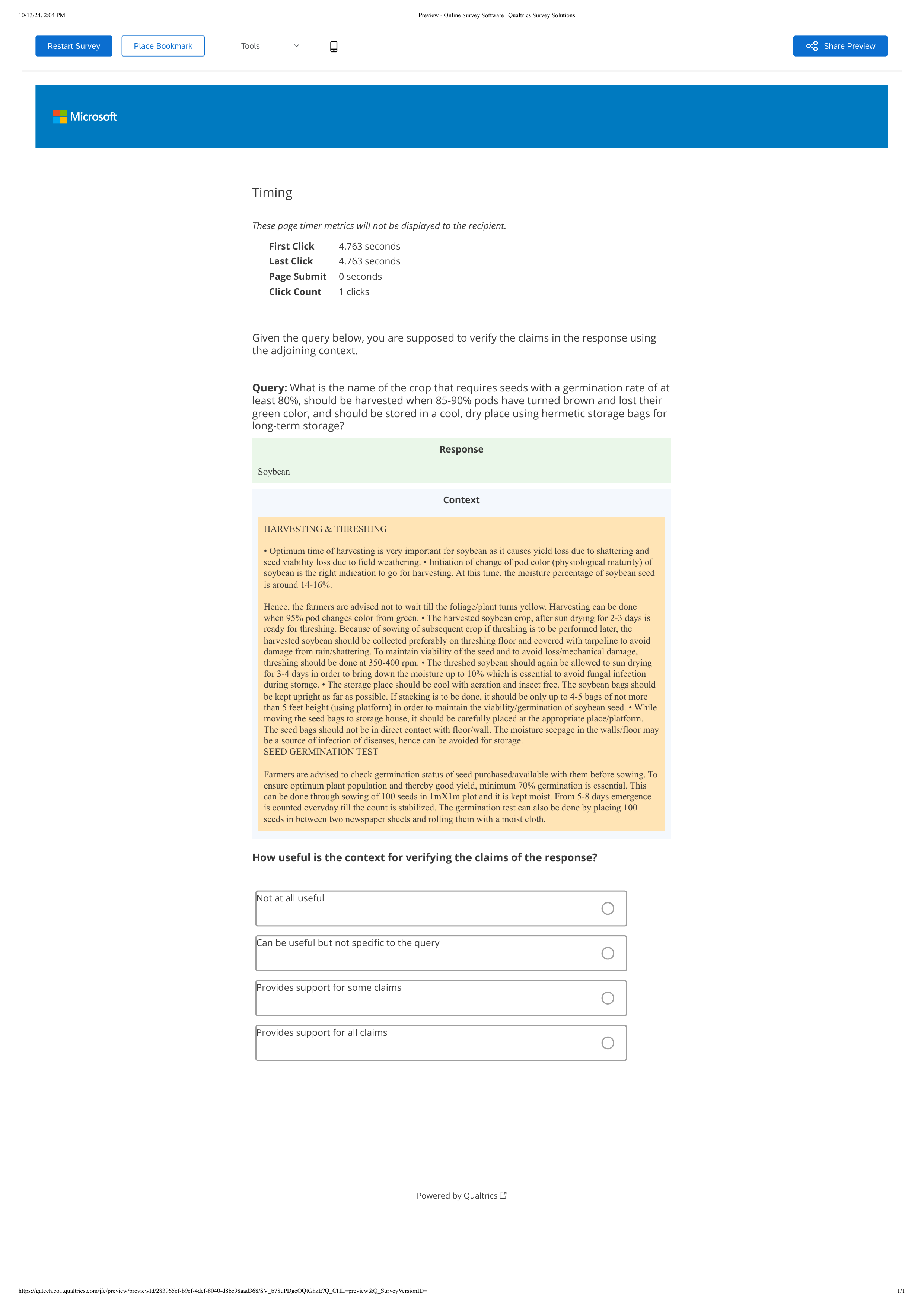}}
    \hfill
    \subfloat[OG-RAG context attribution]{\includegraphics[width=0.3\linewidth, trim={10cm 10cm 10cm 8cm},clip]{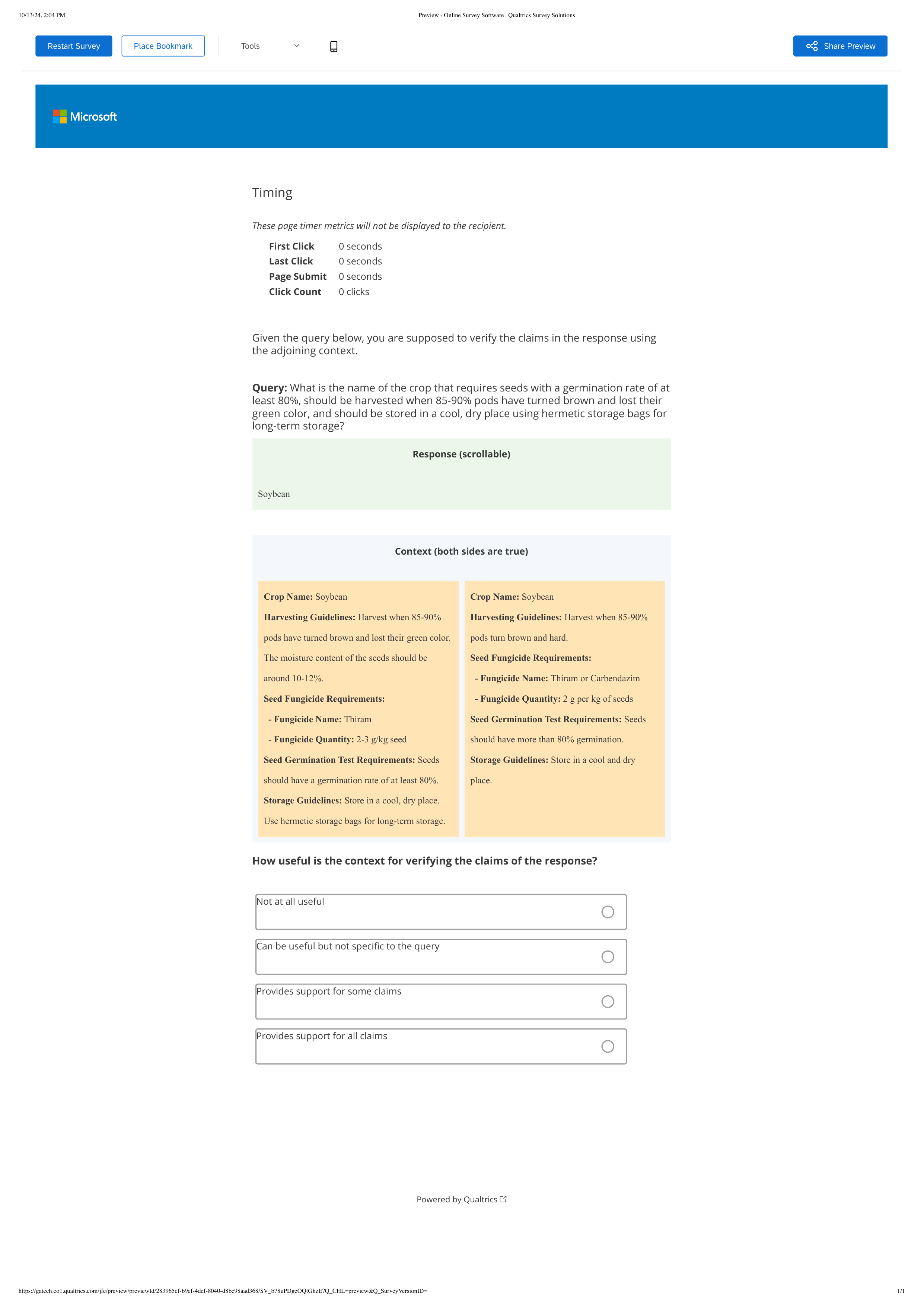}}
    \hspace*{\fill}
    \caption{Survey design}
    \label{fig:survey}
\end{figure*}

Figure~\ref{fig:survey} shows the design of our survey with the introduction and one exemplary question.

\begin{figure*}[!ht]
    \centering
    \input{agri-ontology}
    \caption{Agriculture ontology}
    \label{fig:agri_ontology}
\end{figure*}
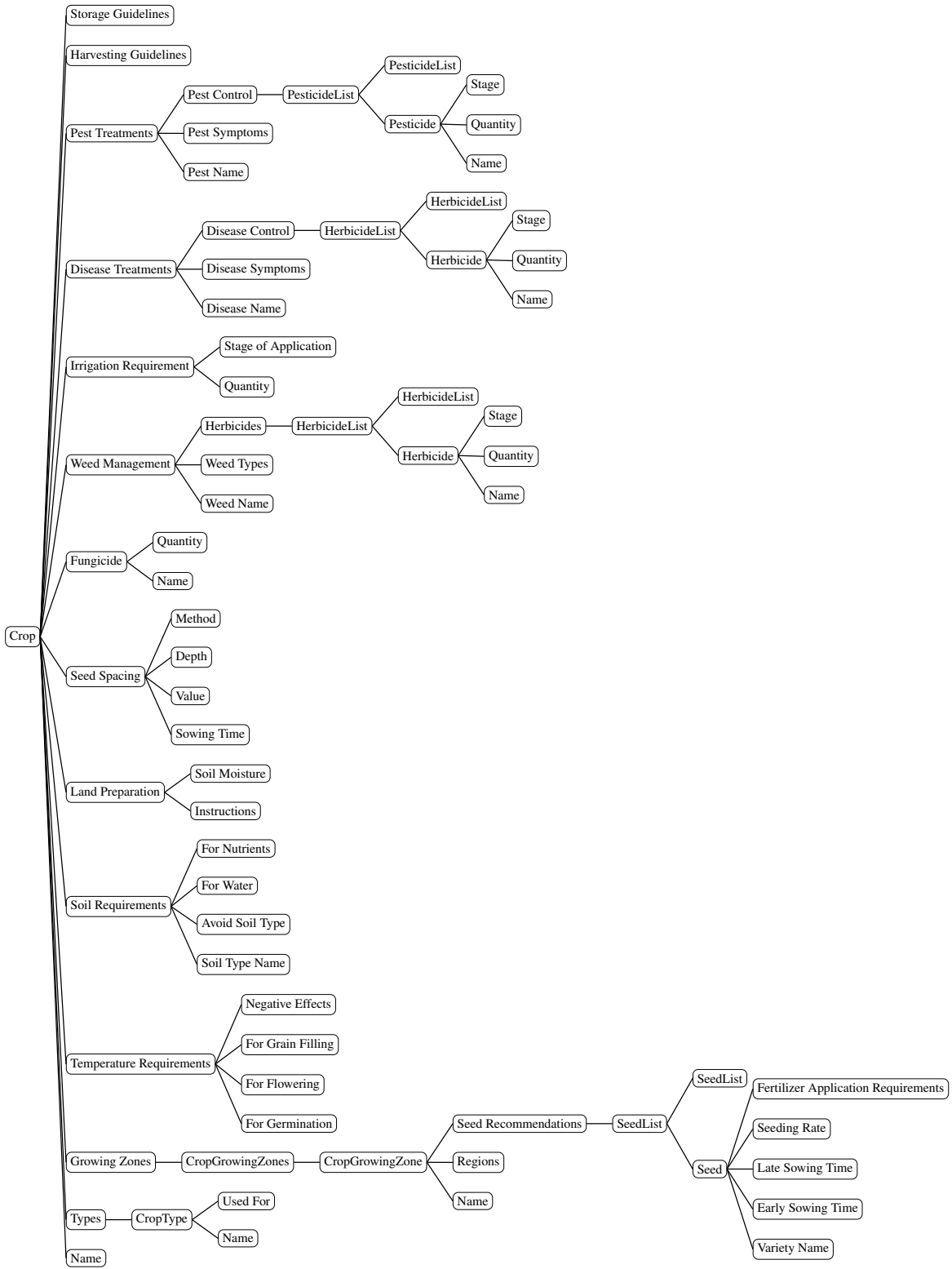

%% file: agri-ontology.tex
\begin{adjustbox}{max size={1.0\textwidth}{0.9\textheight}}
\begin{forest}
for tree={
  grow=east,
  draw,
  rounded corners,
  node options={align=center},
  edge+={thick},
  anchor=west,
  parent anchor=east,
  child anchor=west,
  s sep=15pt,
  l sep=20pt,
}
  [Crop
    [Name]
    [Types
      [CropType
        [Name]
        [Used For]
      ]
    ]
    [Growing Zones
      [CropGrowingZones
        [CropGrowingZone
          [Name]
          [Regions]
          [Seed Recommendations
            [SeedList
                [Seed
                  [Variety Name]
                  [Early Sowing Time]
                  [Late Sowing Time]
                  [Seeding Rate]
                  [Fertilizer Application Requirements]
                ]
                [SeedList]
            ]
          ]
        ]
      ]
    ]
    [Temperature Requirements
      [For Germination]
      [For Flowering]
      [For Grain Filling]
      [Negative Effects]
    ]
    [Soil Requirements
      [Soil Type Name]
      [Avoid Soil Type]
      [For Water]
      [For Nutrients]
    ]
    [Land Preparation
      [Instructions]
      [Soil Moisture]
    ]
    [Seed Spacing
      [Sowing Time]
      [Value]
      [Depth]
      [Method]
    ]
    [Fungicide
      [Name]
      [Quantity]
    ]
    [Weed Management
      [Weed Name]
      [Weed Types]
      [Herbicides
        [HerbicideList
            [Herbicide
              [Name]
              [Quantity]
              [Stage]
            ]
            [HerbicideList]
        ]
      ]
    ]
    [Irrigation Requirement
      [Quantity]
      [Stage of Application]
    ]
    [Disease Treatments
      [Disease Name]
      [Disease Symptoms]
      [Disease Control
        [HerbicideList
            [Herbicide
              [Name]
              [Quantity]
              [Stage]
            ]
            [HerbicideList]
        ]
      ]
    ]
    [Pest Treatments
      [Pest Name]
      [Pest Symptoms]
      [Pest Control
        [PesticideList
          [Pesticide
              [Name]
              [Quantity]
              [Stage]
          ]
          [PesticideList]
        ]
      ]
    ]
    [Harvesting Guidelines]
    [Storage Guidelines]
  ]
\end{forest}
\end{adjustbox}

%% file: news-ontology.tex
\begin{adjustbox}{max size={1.0\textwidth}{0.7\textheight}}
\begin{forest}
for tree={
  grow=south, % Makes the tree grow top to bottom
  draw,
  rounded corners,
  node options={align=center, fill=white}, % Added background color for nodes
  edge+={thick},
  anchor=north,
  parent anchor=south,
  child anchor=north,
  s sep=15pt, % Space between siblings
  l sep=20pt, % Space between levels
}
[NewsAsset
    [Title]
    [Author]
    [Theme]
    [Summary]
    [Created]
    [Has Thumbnail]
    [Embargoed Until]
    [Modified]
    [Published]
    [Has Events
      [EventList
        [Event
          [Event Title]
          [Event Summary]
          [Theme]
          [Has Location]
          [Mentions Persons
            [PersonList
              [Person
                [Date of Birth]
                [Date of Death]
                [Name]
                [Place of Birth]
                [Alias]
              ]
              [PersonList]
            ]
          ]
          [Mentions Organizations
            [OrganizationList
              [Organization
                [Name]
                [Has Location]
                [Alias]
                [Notably Associated With]
              ]
              [OrganizationList]
            ]
          ]
          [Sub Event Of
            [Event]
          ]
          [Notably Associated With
            [Event]
          ]
        ]
        [EventList]
      ]
    ]
]
\end{forest}
\end{adjustbox}